\documentclass{article}

\PassOptionsToPackage{numbers, compress}{natbib}
\usepackage[preprint]{neurips_2026}

\usepackage{amsmath}
\usepackage{graphicx}
\usepackage[table]{xcolor}
\usepackage{multirow}
\usepackage{booktabs}
\usepackage{adjustbox}
\usepackage{float}
\newtheorem{theorem}{Theorem}

\usepackage{amssymb}
\usepackage{hyperref}

\usepackage[utf8]{inputenc} % allow utf-8 input
\usepackage[T1]{fontenc}    % use 8-bit T1 fonts
\usepackage{hyperref}       % hyperlinks
\usepackage{url}            % simple URL typesetting
\usepackage{booktabs}       % professional-quality tables
\usepackage{amsfonts}       % blackboard math symbols
\usepackage{nicefrac}       % compact symbols for 1/2, etc.
\usepackage{microtype}      % microtypography
\usepackage{xcolor}         % colors

\title{CLOVER: Closed-Loop Value Estimation \& Ranking for End-to-End Autonomous Driving Planning}

\author{%
  Sining Ang\textsuperscript{1,2}
  \quad
  Yuguang Yang\textsuperscript{3,2}
  \quad
  Canyu Chen\textsuperscript{4,2}
  \quad
  Yan Wang\textsuperscript{2}\thanks{Corresponding author.}
  \\[2mm]
  \textsuperscript{1}Department of Automation, University of Science and Technology of China
  \\
  \textsuperscript{2}Institute for AI Industry Research, Tsinghua University
  \\
  \textsuperscript{3}School of Electronic Information Engineering, Beihang University
  \\
  \textsuperscript{4}National College for Excellent Engineers, Beihang University
  \\[1mm]
  \texttt{angsn@mail.ustc.edu.cn}
  \quad
  \texttt{wangyan@air.tsinghua.edu.cn}
}

\begin{document}

\maketitle

\begin{abstract}
End-to-end autonomous driving planners are commonly trained by imitating a single logged trajectory, yet they are evaluated by rule-based planning metrics that measure safety, feasibility, progress, and comfort. This creates a training--evaluation mismatch: trajectories close to the logged path may still violate planning rules, while alternative trajectories farther from the demonstration can remain valid and high-scoring. The mismatch is especially limiting for proposal-selection planners, whose performance depends on both candidate-set coverage and scorer ranking quality. We propose \textbf{CLOVER}, a \textbf{C}losed-\textbf{LO}op \textbf{V}alue \textbf{E}stimation and \textbf{R}anking framework for end-to-end autonomous driving planning. CLOVER follows a lightweight generator--scorer formulation: a generator produces diverse candidate trajectories, and a trajectory-level scorer predicts planning-metric sub-scores to rank them at inference time. To expand proposal support beyond single-trajectory imitation, CLOVER constructs evaluator-filtered pseudo-expert trajectories and trains the generator with set-level coverage supervision. It then performs conservative closed-loop self-distillation: the scorer is fitted to true evaluator sub-scores on generated proposals, while the generator is refined toward teacher-selected top-$k$ and vector-Pareto proposal targets with stability regularization. We further analyze when an imperfect scorer can improve the generator, showing that scorer-mediated refinement is reliable when scorer-selected targets are statistically enriched under the true evaluator and updates remain conservative. On NAVSIM, CLOVER achieves 94.5 PDMS and 90.4 EPDMS, establishing a new state-of-the-art performance. On the more challenging NavHard split, it obtains 48.3 EPDMS, matching the strongest reported result. On supplementary nuScenes open-loop evaluation, CLOVER achieves the lowest L2 error and collision rate among compared methods. Code and generated data will be released at \href{https://github.com/WilliamXuanYu/CLOVER}{https://github.com/WilliamXuanYu/CLOVER}.
\end{abstract}

\begin{figure*}[t]
    \centering
    \includegraphics[width=\textwidth]{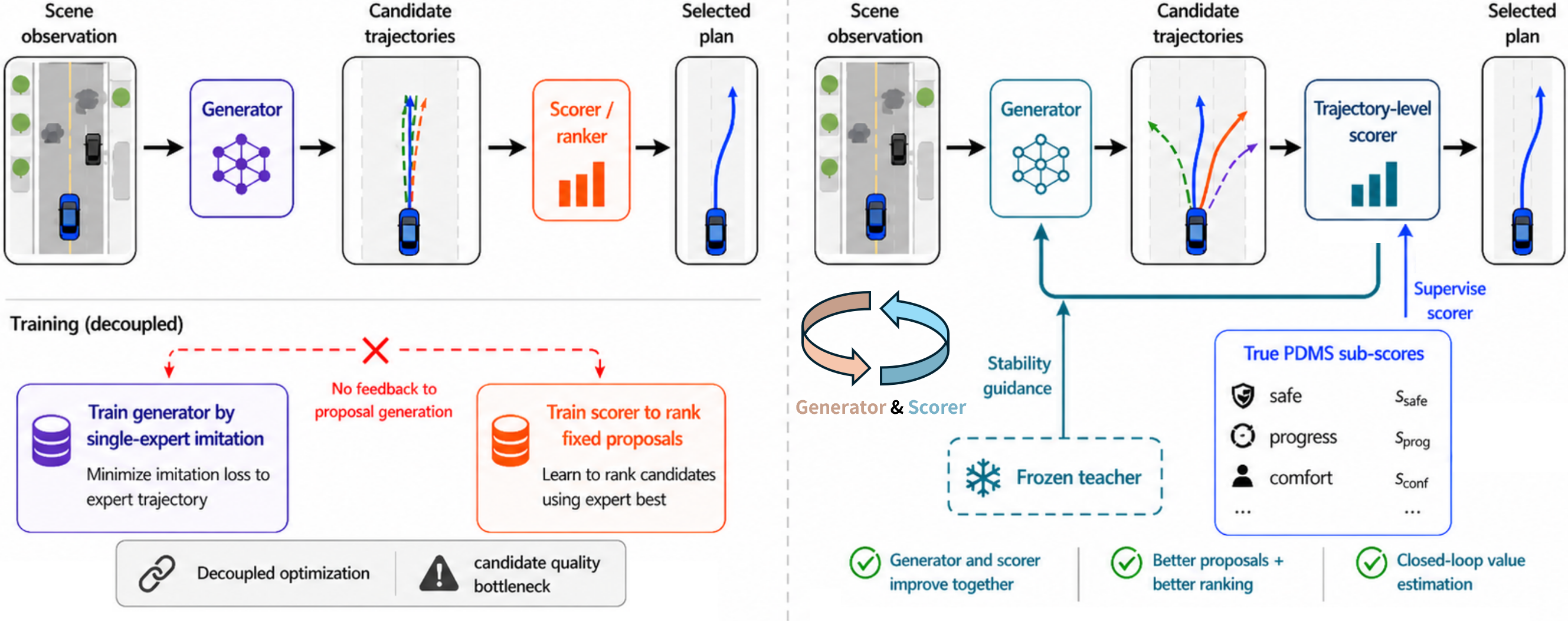}
    \caption{
    \textbf{CLOVER closes the loop between proposal generation and trajectory ranking.}
    \textbf{Left:} Conventional proposal-selection planners generate candidate trajectories and select the final plan with a learned scorer or ranker. During training, however, the generator is typically optimized by single-expert imitation, so ranking feedback does not explicitly reshape the proposal distribution toward higher-value regions.
    \textbf{Right:} CLOVER trains a trajectory-level scorer with true planning-metric sub-scores and conservatively refines the generator through teacher-guided self-distillation. This closed-loop design improves candidate-set coverage and ranking quality together, increasing the chance that high-value trajectories are both generated and selected.
    }
    \label{fig:teaser}
\end{figure*}

\section{Introduction}

End-to-end autonomous driving planning has become a promising paradigm for mapping sensor observations directly to future ego trajectories~\citep{hu2023uniad,jiang2023vad,chitta2022transfuser,guo2025ipad,liao2025diffusiondrive,li2025recogdrive,chen2024vadv2}. 
However, planning is not a standard trajectory-regression problem. 
Most planners are trained to imitate a single logged human trajectory, whereas they are evaluated by rule-based planning metrics such as NAVSIM PDMS/EPDMS~\citep{dauner2024navsim,Cao2025navsim2}, which measure safety, feasibility, progress, and comfort. 
A trajectory close to the logged path may still collide, leave the drivable area, or induce unsafe interactions, while another trajectory farther from the demonstration may remain valid and high-scoring. 
This training--evaluation mismatch limits purely imitation-based planning under closed-loop-style metrics.

The mismatch is particularly important for proposal-selection planners. 
These methods first generate multiple candidate trajectories and then select one using a learned scorer or ranker. 
Their final performance is constrained by two coupled factors: whether the generator covers high-quality alternatives, and whether the scorer can identify them. 
Single-trajectory supervision can collapse candidates around one realized behavior, even when multiple decisions are valid in the same scene, such as maintaining speed, braking, or choosing a different interaction strategy. 
Conversely, even a strong candidate set can fail if the scorer ranks unsafe or infeasible trajectories above better alternatives. 
Thus, effective proposal-selection planning requires both broad proposal coverage and reliable use of scorer feedback.

We propose \textbf{CLOVER}, a \textbf{C}losed-\textbf{LO}op \textbf{V}alue \textbf{E}stimation and \textbf{R}anking framework for proposal-selection based end-to-end autonomous driving planning. 
Given multi-view camera observations and ego state, CLOVER uses a lightweight generator--scorer architecture: the generator produces candidate future ego trajectories, and a trajectory-level scorer predicts planning-metric sub-scores for each candidate. 
At inference time, candidates are ranked by the composed predicted score and the top-ranked trajectory is selected. 
During training, the true rule-based evaluator supervises scorer learning and generator refinement, but the evaluator is not required at deployment.

CLOVER addresses the training--evaluation mismatch in two stages. 
First, it expands single-trajectory imitation into \textbf{set-level proposal coverage}. 
Evaluator-filtered pseudo-expert trajectories are constructed from diverse trajectory families, including different lateral offsets, speed profiles, and boundary behaviors, and the generator is trained to cover them while retaining the logged trajectory as a driving prior. 
This exposes the generator to multiple feasible driving modes and raises the oracle upper bound of the proposal set. 
Second, CLOVER performs \textbf{conservative closed-loop self-distillation}. 
The scorer is fitted to true evaluator sub-scores on generated proposals; a frozen teacher then provides top-$k$ and vector-Pareto proposal targets, which the student generator covers with stability regularization. 
This lets the scorer guide generator improvement without unrestricted maximization of an imperfect learned reward.

A key question is when such scorer-mediated refinement is trustworthy. 
CLOVER does not require a globally perfect scorer. 
Our analysis shows that it is sufficient for scorer-selected targets to be statistically better under the true evaluator than the current proposal distribution. 
Under this selected-set enrichment condition, conservative set-level distillation increases the generator's probability mass on high-quality trajectories. 
This provides a direct condition under which an imperfect learned scorer can serve as a useful mediator for proposal refinement. 
We further validate this condition empirically, showing that scorer-selected targets are enriched under the true evaluator and that Stage-2 refinement improves proposal quality without collapsing diversity.

We evaluate CLOVER on NAVSIM v1, NAVSIM v2 EPDMS, NavHard, and supplementary nuScenes open-loop evaluation. 
CLOVER matches or surpasses the strongest reported results across the main NAVSIM-style benchmarks, achieving 94.5 PDMS and 90.4 EPDMS on NAVSIM and 48.3 EPDMS on NavHard, while also obtaining strong nuScenes open-loop results. 
Further analyses show that CLOVER improves both proposal quality and proposal diversity, indicating that its gains come from a stronger candidate distribution rather than only from selecting a better top-1 trajectory. 
Although CLOVER superficially resembles actor--critic methods, it does not interact with the environment or perform trial-and-error exploration; all feedback is computed offline from logged scenes and a rule-based evaluator.

Our contributions are summarized as follows:
\begin{itemize}
    \item We propose \textbf{CLOVER}, a closed-loop value estimation and ranking framework that couples proposal generation and trajectory scoring.

    \item We introduce evaluator-filtered pseudo-expert coverage supervision, extending single-trajectory imitation to set-level multi-modal proposal training and improving candidate-set quality.

    \item We develop conservative closed-loop self-distillation, alternating between scorer fitting on true planning sub-scores and teacher-guided generator refinement with top-$k$, vector-Pareto, and stability objectives.

    \item We analyze when imperfect-scorer guidance is reliable under selected-set enrichment, and empirically show that CLOVER improves proposal quality while preserving diversity across NAVSIM-style evaluations.
\end{itemize}

\section{Related Work}

\subsection{End-to-end Autonomous Driving Planning}

End-to-end autonomous driving planning aims to predict future ego trajectories directly from sensor observations. 
Early methods typically mapped front-view images to low-level control commands, while recent approaches use multi-view encoders, Transformer decoders, BEV representations, trajectory queries, and vision pretraining to unify scene understanding and planning~\citep{hu2023uniad,jiang2023vad,chitta2022transfuser,guo2025ipad,liao2025diffusiondrive,li2025recogdrive,chen2024vadv2}. 
These methods have significantly improved planning performance, especially under modern evaluation protocols such as NAVSIM~\citep{dauner2024navsim,Cao2025navsim2}. 
However, most end-to-end planners are still trained primarily by imitating logged expert trajectories. 
Such supervision is stable and data-efficient, but it is not fully aligned with rule-based planning metrics that evaluate safety, feasibility, progress, and comfort. 
CLOVER differs by explicitly using evaluator-provided planning feedback to train a trajectory-level value estimator and to refine the candidate proposal distribution, rather than relying only on single-trajectory imitation.

\subsection{Multimodal Trajectory Generation and Proposal-selection Planning}

Driving behavior is inherently multi-modal: in the same scene, multiple future trajectories, such as maintaining speed, yielding, braking, or choosing a slightly different interaction strategy, may all be valid. 
Proposal-selection planners address this by generating a set of candidate trajectories and selecting one for execution. 
One line of work uses discretized trajectory vocabularies or anchor sets, such as Hydra-MDP~\citep{li2024hydra} and VADv2~\citep{chen2024vadv2}, which provide a structured candidate space but are constrained by the coverage of the predefined vocabulary. 
Another line uses generative models, such as DiffusionDrive~\citep{liao2025diffusiondrive} and GoalFlow~\citep{xing2025goalflow}, to produce diverse trajectories with greater flexibility, often at higher computational cost. 
Proposal-centric planners such as iPad~\citep{guo2025ipad} refine candidate trajectories online and offer an efficient alternative. 
Despite these advances, proposal generation is often still supervised mainly by the single logged trajectory, which can limit candidate-set coverage when multiple high-quality behaviors are possible. 
CLOVER addresses this limitation by constructing evaluator-filtered pseudo-expert trajectories and training the generator with set-level coverage supervision, thereby improving both proposal diversity and candidate-set quality.

\subsection{Learned Trajectory Scoring}

A key component of proposal-selection planning is trajectory scoring: given a candidate set, the planner must identify the safest and most effective trajectory. 
Learned cost functions, ranking modules, and trajectory scorers have been used to improve selection quality, especially when paired with rich proposal sets~\citep{yao2025drivesuprim,kirby2026driving,li2025generalized}. 
However, trajectory scoring in autonomous driving is challenging because the true planning score depends on map constraints, future obstacle states, interaction rules, and non-differentiable metric computations. 
A learned scorer must estimate this quality from the current observation and candidate trajectory, and directly maximizing such an imperfect score can exploit scorer errors or reduce diversity. 
CLOVER treats the scorer as a closed-loop value estimator trained on evaluator-provided sub-scores, and uses it conservatively: instead of unrestricted score maximization, the generator is refined through teacher-selected top-$k$ and vector-Pareto proposal targets with stability regularization.

\subsection{Relation to Reinforcement Learning}

CLOVER shares a superficial similarity with actor--critic reinforcement learning: a generator proposes trajectories and a scorer evaluates them. 
However, CLOVER is not a reinforcement learning method. 
Classic reinforcement learning relies on environment interaction, trial-and-error exploration, and reward optimization over visited states. 
In contrast, CLOVER operates offline on logged driving scenes; all feedback is computed from generated candidate trajectories using a rule-based planning evaluator, and no online environment interaction or exploration is performed. 
Recent works such as Evadrive~\citep{jiao2025evadrive} explore RL-style planning for autonomous driving, whereas CLOVER focuses on closed-loop value estimation and conservative self-distillation for proposal-selection planning. 
Thus, CLOVER is better viewed as an offline evaluator-guided ranking and proposal-refinement framework rather than an RL-based planner.

\section{Method}
\label{sec:method}

\begin{figure}[t]
    \centering
    \includegraphics[width=\linewidth]{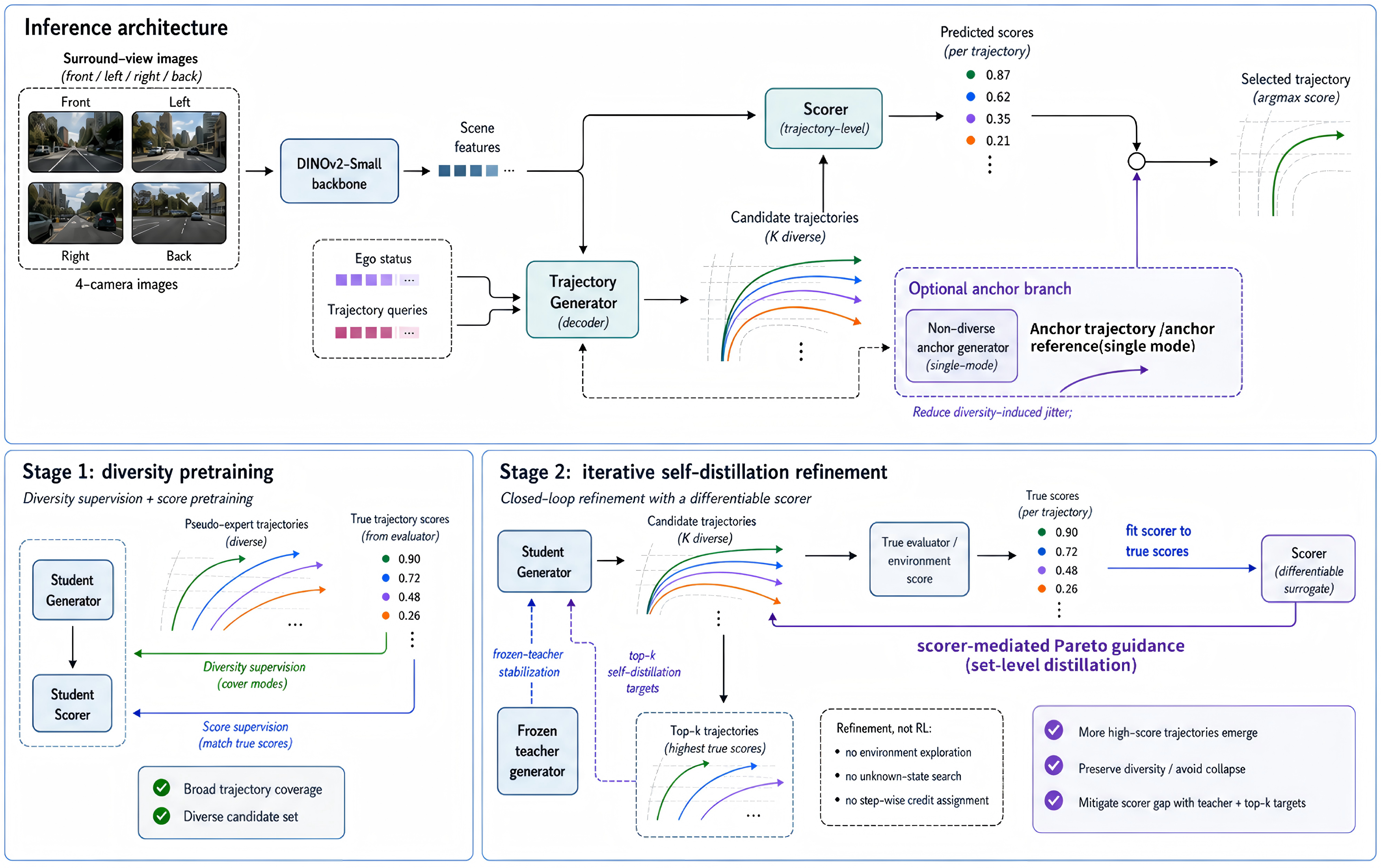}
    \caption{
    \textbf{Overview of CLOVER.}
    At inference time, multi-view images and ego state are encoded into scene features.
    A generator produces $K$ candidate trajectories, and a trajectory-level scorer ranks them by predicted planning-metric sub-scores.
    Stage 1 expands proposal coverage with evaluator-filtered pseudo-expert trajectories.
    Stage 2 fits the scorer to evaluator-provided sub-scores and refines the generator through teacher-guided top-$k$ and vector-Pareto distillation with stability regularization.
    An optional anchor-assisted soft reranking module can be used for EPDMS evaluation to reduce diversity-induced jitter among near-tied candidates.
    }
    \label{fig:clover_overview}
\end{figure}

\textbf{Overview.}
Figure~\ref{fig:clover_overview} summarizes CLOVER.
CLOVER follows a proposal-selection formulation: a generator produces a set of candidate ego trajectories, and a trajectory-level scorer ranks them by predicted planning-metric sub-scores.
Training has two stages.
Stage 1 expands proposal support with evaluator-filtered pseudo-expert coverage supervision and scorer pretraining.
Stage 2 performs conservative closed-loop self-distillation by alternating scorer fitting on evaluator-provided sub-scores and teacher-guided generator refinement.

\subsection{Proposal Selection and Scoring}
\label{sec:proposal_scoring}

Given a driving observation $o$, including multi-view camera images and ego-state information, CLOVER predicts future ego trajectories in the ego-centric frame.
Each trajectory is represented as
$\tau=\{(x_t,y_t,\theta_t)\}_{t=1}^{T}$, where $(x_t,y_t,\theta_t)$ denotes the ego pose at future step $t$.
In our implementation, the prediction horizon is 4 seconds with a 0.5-second interval, resulting in $T=8$ future poses.

Instead of outputting a single trajectory, the generator produces $K$ candidates
\begin{equation}
    \mathcal{T}_{\theta}(o)=G_{\theta}(o)=\{\tau_i\}_{i=1}^{K}.
\end{equation}
A trajectory-level scorer predicts planning-metric sub-scores for each candidate,
\begin{equation}
    \hat{\mathbf r}_i=S_{\phi}(o,\tau_i),
\end{equation}
covering components such as collision avoidance, drivable-area compliance, time-to-collision, progress, and comfort.
These predicted sub-scores are composed by the official PDMS rule $\Gamma_{\mathrm{PDMS}}(\cdot)$, and inference selects
\begin{equation}
    \tau^\ast
    =
    \arg\max_{\tau_i\in\mathcal{T}_{\theta}(o)}
    \Gamma_{\mathrm{PDMS}}(S_{\phi}(o,\tau_i)).
\end{equation}

During training and analysis, the rule-based evaluator $E$ assigns true targets
$\mathbf r_i^{\mathrm{true}}=E(o,\tau_i)$.
Since $E$ involves map queries, future-obstacle states, geometric checks, and discrete rules, it is non-differentiable with respect to the generator.
CLOVER uses the evaluator only to supervise the scorer and construct distillation targets during training; it is not required at deployment.
Exact PDMS/EPDMS definitions are provided in Appendix~\ref{app:offline_score}.

We instantiate this formulation with a lightweight DrivoR-style generator--scorer architecture~\citep{kirby2026driving}.
A DINOv2-Small visual encoder with LoRA fine-tuning maps 4-camera images to compact scene tokens, which are combined with ego-state features and decoded by a Transformer trajectory generator with learnable trajectory queries.
The scorer embeds each candidate trajectory through its explicit waypoints, attends to the same scene tokens, and predicts planning-metric sub-scores.
Thus, the generator determines which behaviors are available for selection, while the scorer estimates their closed-loop trajectory quality.
We use $K=64$ candidates unless otherwise specified; implementation details are in Appendix~\ref{app:implementation_details}.

\subsection{Stage 1: Evaluator-filtered Proposal Coverage}
\label{sec:pseudo_expert}

Single-trajectory imitation treats the logged future trajectory as the only positive target, which is insufficient for training a high-coverage proposal generator.
CLOVER therefore constructs a training-time pseudo-expert set $\mathcal{E}(o)$ for each scene.
These pseudo-experts are not simple perturbations of the logged trajectory.
Instead, they are generated from interpretable candidate families using privileged metric-cache information available during training, including route centerlines, drivable-area maps, and future obstacle occupancy.
The candidate pool covers diverse lateral offsets, acceleration and deceleration profiles, stop-go behaviors, approach-brake trajectories, boundary cases, and overshoot-style motions.
Candidates are pre-filtered by lightweight drivable-area and occupancy checks, scored by the evaluator, and selected with a coverage-aware strategy over score patterns and bins.

During training, we further keep high-quality candidates above a score threshold and apply greedy farthest-point sampling in trajectory space to obtain at most $M$ pseudo-experts,
$\mathcal{E}(o)=\{e_j\}_{j=1}^{M}$.
This set is used only during training and is never required at inference time.
Detailed generation rules and qualitative examples are provided in Appendix~\ref{app:pseudo_expert_generation}.

Given proposals $\mathcal{T}_{\theta}(o)=\{\tau_i\}_{i=1}^{K}$ and the logged trajectory $\tau^{\mathrm{gt}}$, Stage 1 optimizes
{\small
\begin{equation}
\begin{aligned}
    \mathcal{L}_{\mathrm{stage1}}
    &=
    \lambda_{\mathrm{gt}}
    \underbrace{\min_i \|\tau_i-\tau^{\mathrm{gt}}\|_1}_{\mathcal{L}_{\mathrm{gt}}}
    +
    \lambda_{\mathrm{pe}}
    \underbrace{
    \frac{1}{|\mathcal{E}(o)|}
    \sum_{e\in\mathcal{E}(o)}
    \min_i \|\tau_i-e\|_1
    }_{\mathcal{L}_{\mathrm{pe}}} \\
    &\quad +
    \lambda_{\mathrm{score}}
    \underbrace{
    \frac{1}{|\mathcal{Q}(o)|}
    \sum_{\tau\in\mathcal{Q}(o)}
    \ell\big(S_{\phi}(o,\tau),E(o,\tau)\big)
    }_{\mathcal{L}_{\mathrm{score}}}.
\end{aligned}
\end{equation}
}
Here, $\mathcal{L}_{\mathrm{gt}}$ preserves the logged driving prior, $\mathcal{L}_{\mathrm{pe}}$ encourages the proposal set to cover multiple pseudo-expert modes, and $\mathcal{L}_{\mathrm{score}}$ pretrains the scorer to predict evaluator-provided sub-scores.
The loss $\ell$ uses component-specific classification or regression targets.
This stage expands the candidate distribution beyond a single realized human trajectory while retaining stable imitation supervision.

\subsection{Stage 2: Conservative Closed-loop Self-distillation}
\label{sec:stage2}

Stage 2 refines the expanded proposal distribution using evaluator feedback.
Directly maximizing a learned scalar score can exploit scorer errors or collapse diversity, so CLOVER uses an alternating procedure initialized from Stage 1: a scorer phase fits the value estimator on current proposals, and a generator phase distills high-value teacher proposals under stability regularization.

\textbf{Scorer phase.}
The current generator produces $\mathcal{T}_{\theta}(o)=\{\tau_i\}_{i=1}^{K}$, and the evaluator provides true sub-score targets $\mathbf r_i^{\mathrm{true}}=E(o,\tau_i)$.
The scorer minimizes
\begin{equation}
    \mathcal{L}_{\mathrm{critic}}
    =
    \frac{1}{K}
    \sum_{i=1}^{K}
    \ell(S_{\phi}(o,\tau_i),\mathbf r_i^{\mathrm{true}}),
\end{equation}
distilling the non-differentiable evaluator into a differentiable trajectory-level value estimator on the current proposal distribution.

\textbf{Generator phase.}
A frozen teacher generates
$\mathcal{T}^{\mathrm{tea}}(o)=\{\tau_i^{\mathrm{tea}}\}_{i=1}^{K}$.
The scorer evaluates these proposals and constructs two target sets.
The first is the top-$k$ set $\mathcal{K}^{\mathrm{tea}}$, selected by the composed predicted score $\Gamma_{\mathrm{PDMS}}(S_{\phi}(o,\tau))$.
The second is the vector-Pareto set $\mathcal{P}^{\mathrm{tea}}$, consisting of non-dominated proposals in predicted sub-score space.
A trajectory $\tau_a$ dominates $\tau_b$ if $r_a^m\ge r_b^m$ for all components $m$ and $r_a^m>r_b^m$ for at least one component.

The student proposal set $\mathcal{T}^{\mathrm{stu}}(o)=\{\tau_i^{\mathrm{stu}}\}_{i=1}^{K}$ is trained to cover these targets with
{\small
\begin{equation}
    \mathcal{L}_{\mathrm{set}}(\mathcal{T}^{\mathrm{stu}},\mathcal{A})
    =
    \frac{1}{|\mathcal{A}|}
    \sum_{\tau\in\mathcal{A}}
    \min_i \|\tau_i^{\mathrm{stu}}-\tau\|_1,
    \quad
    \mathcal{A}\in\{\mathcal{K}^{\mathrm{tea}},\mathcal{P}^{\mathrm{tea}}\}.
\end{equation}
}
The generator objective is
{\small
\begin{equation}
\begin{aligned}
    \mathcal{L}_{\mathrm{gen}}
    &=
    \lambda_{\mathrm{traj}}\mathcal{L}_{\mathrm{gt}}
    +
    \lambda_{\mathrm{topk}}
    \mathcal{L}_{\mathrm{set}}(\mathcal{T}^{\mathrm{stu}},\mathcal{K}^{\mathrm{tea}})
    +
    \lambda_{\mathrm{pareto}}
    \mathcal{L}_{\mathrm{set}}(\mathcal{T}^{\mathrm{stu}},\mathcal{P}^{\mathrm{tea}}) \\
    &\quad +
    \lambda_{\mathrm{stab}}
    \frac{1}{K}
    \sum_{i=1}^{K}
    \|\tau_i^{\mathrm{stu}}-\operatorname{sg}(\tau_i^{\mathrm{tea}})\|_1 .
\end{aligned}
\end{equation}
}
The top-$k$ term concentrates probability mass on high-scoring proposals, the vector-Pareto term preserves safety--progress--comfort trade-offs, and the stability term keeps refinement conservative by discouraging drift away from the teacher proposal distribution.

\subsection{Inference and Optional Anchor Reranking}
\label{sec:inference_anchor}

At inference time, CLOVER uses only the generator and learned scorer: candidates are generated, scored, and ranked by the composed predicted metric score.
The evaluator, pseudo-expert generator, and self-distillation procedure are used only during training and analysis.

For NAVSIM v2 EPDMS, we optionally apply anchor-assisted soft reranking to reduce diversity-induced jitter among near-tied high-quality candidates.
This is useful because EPDMS includes extended-comfort terms that depend on cross-frame decision consistency, whereas the scorer predicts per-trajectory planning-metric sub-scores.
The anchor is produced by a single-mode planner trained with the same architecture and data protocol but with diversity-related settings disabled and only standard single-trajectory imitation retained.
It uses no future information or evaluator scores at inference time; it only provides a stable reference trajectory.
Candidates are reranked by combining the learned scorer score with a continuous penalty for position and heading deviation from the anchor.
This soft reranking does not discard candidates with hard thresholds; it only biases selection toward temporally stable choices among near-tied high-quality proposals.
The exact formula and parameter sweep are provided in Appendix~\ref{app:anchor_soft}.

\section{Theoretical Analysis}
\label{sec:theory}

We provide a concise theoretical justification for when CLOVER's scorer-mediated refinement can improve the generator.
The key question is not whether the learned scorer is globally perfect, but whether the targets selected by the scorer are statistically better under the true evaluator than the current proposal distribution.
We therefore focus on a selected-set enrichment condition: if the scorer-selected top-$k$ or vector-Pareto targets contain more true high-quality trajectories than the current proposal set, then conservative set-level distillation increases the generator's support on high-quality trajectories.
Full proofs, auxiliary results on approximate monotonicity, scorer refitting, and Pareto consistency, and additional empirical validation are provided in Appendix~\ref{app:theory}.

All scores are normalized to $[0,1]$.
For a scene $o$, let
\[
    \mu_t^o
    =
    \frac{1}{K}
    \sum_{i=1}^{K}
    \delta_{\tau_i^t(o)}
\]
denote the empirical distribution of the current $K$ proposals.
Let $A_t(o)$ be the scorer-selected target set, such as the top-$k$ or vector-Pareto teacher proposals, and let $\nu_t^o$ be the empirical distribution over $A_t(o)$.
For a high-score region
\[
    \mathcal{H}_o
    =
    \{\tau:R^*(o,\tau)\ge r_{\mathrm{high}}\},
\]
define
$p_t(o)=\mu_t^o(\mathcal{H}_o)$ and
$q_t(o)=\nu_t^o(\mathcal{H}_o)$.

\begin{theorem}[Selected-set enrichment improves high-score support]
\label{thm:enrichment}
Assume that the scorer-selected target set is enriched with true high-score trajectories:
\[
    q_t(o)\ge p_t(o)+\xi_t(o),
    \qquad
    \xi_t(o)>0.
\]
Assume that the generator update is conservative, i.e.,
\[
    \mathrm{TV}\!\left(
    \mu_{t+1}^o,
    (1-\alpha_t)\mu_t^o+\alpha_t\nu_t^o
    \right)
    \le \eta_t(o),
    \qquad
    \alpha_t\in[0,1].
\]
Then
\[
    p_{t+1}(o)
    \ge
    p_t(o)+\alpha_t\xi_t(o)-\eta_t(o).
\]
Therefore, if $\alpha_t\xi_t(o)>\eta_t(o)$, Stage-2 refinement increases the probability mass assigned to high-score trajectories.

Similarly, if the selected targets have higher true expected score,
\[
    \mathbb{E}_{\nu_t^o}[R^*]
    -
    \mathbb{E}_{\mu_t^o}[R^*]
    \ge
    \beta_t(o)>0,
\]
then
\[
    \mathbb{E}_{\mu_{t+1}^o}[R^*]
    \ge
    \mathbb{E}_{\mu_t^o}[R^*]
    +
    \alpha_t\beta_t(o)
    -
    \eta_t(o).
\]
\end{theorem}

\paragraph{Implication for CLOVER.}
The theorem states a weak and testable condition: the scorer does not need to rank every trajectory correctly, nor does it need to be accurate over the entire trajectory space.
It only needs to select target sets that are statistically enriched with true high-quality trajectories under the evaluator.
This matches CLOVER's design.
Stage 1 provides broad proposal support, the scorer phase calibrates the value estimator on generated proposals, and the generator phase conservatively distills scorer-selected top-$k$ and vector-Pareto targets rather than directly maximizing a learned scalar reward.
The stability term controls the deviation from the teacher proposal distribution, corresponding to the conservative-update error $\eta_t(o)$.

\paragraph{Empirical support.}
We empirically validate the two premises behind Theorem~\ref{thm:enrichment}: the proposal pool must already contain high-quality modes, and the scorer-selected targets must be enriched under the true evaluator.
On 12,146 NAVSIM scenes with 64 proposals per scene, the pooled proposal distribution contains substantial high-score support, with $35.42\%$ full-score proposals.
More importantly, high-confidence scorer-selected proposals are much more likely to be truly high scoring: among proposals with predicted score $s\ge0.95$, the true mean score is $0.9753$, $P(R^*\ge0.90)=96.69\%$, and $P(R^*=1)=69.74\%$.
Thus, for the full-score region $\mathcal{H}_o=\{\tau:R^*(o,\tau)=1\}$, the empirical enrichment gap is approximately
$69.74\%-35.42\%=34.32\%$, directly matching the condition
$q_t(o)\ge p_t(o)+\xi_t(o)$.

We further evaluate whether the scorer is useful for target selection, rather than globally perfect trajectory scoring.
For fixed groups with true score $\ge0.95$ versus $\le0.50$, the scorer reaches $94.72\%$ pairwise ranking accuracy, and the elite/reject margin condition $\gamma_o>2\epsilon_{p90,o}$ holds in $82.37\%$ of scenes.
Finally, when the scorer is used to select target sets from the proposal pool, the scorer-selected Top-8 and Top-16 sets achieve mean best true scores of $0.9656$ and $0.9730$, respectively, supporting the expected-score version of Theorem~\ref{thm:enrichment}.
These diagnostics are not used as final benchmark results; rather, they verify that the target-selection step provides evaluator-enriched supervision for Stage-2 distillation.
Overall, Stage 2 does not rely on a globally perfect scorer or on discovering high-quality modes from scratch.
Instead, it reallocates generator probability mass toward evaluator-enriched targets selected from an already broad proposal space.
Additional empirical validation is provided in Appendix~\ref{app:theory}.

\section{Experiments}
\label{sec:experiments}

\subsection{Experimental Setup}
\label{sec:experimental_setup}

We evaluate CLOVER on NAVSIM v1, NAVSIM v2 EPDMS, NavHard, and supplementary nuScenes open-loop planning.
Our main evaluation follows NAVSIM-style planning metrics, where trajectory quality is measured by rule-based safety, progress, and comfort components rather than only displacement error to the logged trajectory.
NAVSIM v1 reports PDMS, while NAVSIM v2 reports EPDMS on the \texttt{navtest} split~\citep{dauner2024navsim,Cao2025navsim2}.
We further evaluate on the more challenging \texttt{navhard-two-stage} split using EPDMS.
For supplementary open-loop validation on nuScenes~\citep{nuscenes}, we follow the ST-P3 and UniAD protocols and report L2 displacement error and collision rate~\citep{hu2022st,hu2023planning}.
Exact PDMS/EPDMS definitions are provided in Appendix~\ref{app:offline_score}.

Unless otherwise specified, all experiments use the same DrivoR-style generator--scorer architecture~\citep{kirby2026driving}.
The model uses 4-camera images without LiDAR, a DINOv2 ViT-S visual encoder with LoRA fine-tuning~\citep{oquab2024dinov2,hu2022lora}, 64 trajectory proposals, and a 4-second horizon sampled every 0.5 seconds.
Stage 1 is trained with pseudo-expert coverage and scorer pretraining, and Stage 2 is initialized from Stage 1 and trained with alternating scorer-fitting and generator-refinement phases.
Because CLOVER modifies training but keeps the deployed generator--scorer inference graph unchanged, it introduces no additional inference-time modules over the base DrivoR-style architecture.
With $K=64$ proposals, the deployed model runs at about 110 ms per scene on a single NVIDIA A100 GPU in the NAVSIM-v2 setting.
Detailed architecture, hyperparameters, and training cost are provided in Appendix~\ref{app:implementation_details}.

\subsection{Main Results}
\label{sec:main_results}

Tables~\ref{tab:benchmark_navsim_v1_small}--\ref{tab:nuscenes_l2} summarize results on NAVSIM v1, NAVSIM v2 \texttt{navtest}, NavHard, and supplementary nuScenes open-loop evaluation.
CLOVER achieves \textbf{94.5 PDMS} on NAVSIM v1, improving over the strongest generator--scorer baseline DrivoR~\citep{kirby2026driving} by 0.8 points and ranking first or second across all PDMS sub-scores.
Under NAVSIM v2, it obtains \textbf{90.4 EPDMS} with the updated official implementation and \textbf{87.2 EPDMS$^\ast$} with the original evaluation code.
On the challenging \texttt{navhard-two-stage} split, CLOVER reaches \textbf{48.3 EPDMS}, matching the strongest reported result.
It also achieves strong supplementary nuScenes open-loop results, including the lowest ST-P3 L2 error and best or tied-best collision rate.

We also repeat the main CLOVER training with three random seeds and observe negligible variation: all runs differ by less than $0.02$ PDMS and round to the same one-decimal benchmark score; detailed seed-stability results are provided in Appendix~\ref{app:seed_stability}.

\begin{table}[t]
    \centering
    \caption{
    \textbf{NAVSIM v1 benchmark.}
    Comparison with published and recent methods under the official PDMS metric. 
    Higher is better for all columns. 
    CLOVER achieves the best PDMS and ranks first or second across all sub-scores, nearly matching the human-driver reference.
    }
    \small
     \setlength{\tabcolsep}{1.8pt}
    % \resizebox{\columnwidth}{!}{
    \begin{tabular}{@{}l@{}rl|ccccc|c@{}}
    \toprule
    Method & & & NC & DAC & TTC & Comf.& EP & \textbf{PDMS} \\
    \midrule
    \rowcolor{black!10}
    PDMS‑Closed & \cite{dauner2023pdm}               & \scriptsize{PMLR'23} & 94.6 & 99.8 & 89.9 & 86.9 & 99.9 & 89.1 \\ %& Published\\
    \rowcolor{black!10}
    Human driver& \cite{dauner2024navsim}           & \scriptsize{NeurIPS'24} & 100 & 100&  100 & 99.9 & 87.5 & 94.8 \\ %& Published \\

    \midrule
    Ego‑stat. MLP & \cite{dauner2024navsim}        & \scriptsize{NeurIPS'24} & 93.0 & 77.3 & 83.6 & 100 & 62.8 & 65.6 \\ %& Published	\\
    UniVLA  & \cite{wang2025unified}                & \scriptsize{arXiv'25} & 96.9 & 91.1 & 91.7 & 96.7 & 76.8 & 81.7 \\ %& Arxiv (2025/06)\\
    DrivingGPT & \cite{chen2024drivinggpt}          & \scriptsize{ICCV'24} & 98.9 & 90.7 & 94.9 & 95.6 & 79.7 & 82.4 \\ %& Arxiv (2024/12)\\
    UniAD & \cite{hu2023uniad}                   & \scriptsize{CVPR'23} & 97.8 & 91.9 & 92.9 & 100  & 78.8 & 83.4 \\ %& Published\\
    LTF & \cite{chitta2022transfuser}               & \scriptsize{TPAMI'22} & 97.4 & 92.8 & 92.4 & 100  & 79.0 & 83.8 \\ %& Published \\
    PARA‑Drive & \cite{paradrive}               & \scriptsize{CVPR'24} & 97.9 & 92.4 & 93.0 & 99.8 & 79.3 & 84.0 \\ %& Published \\
    DriveX-S & \cite{shi2025drivex}                 & \scriptsize{ICCV'25} & 97.5 & 94.0 & 93.0 & 100  & 79.7 & 84.5 \\ %& Arxiv (2025/05)\\
    World4Drive & \cite{zheng2025world4drive}       & \scriptsize{ICCV'25} & 97.4 & 94.3 & 92.8 & 100  & 79.9 & 85.1 \\ %& Arxiv (2025/07) (ICLR26 submission) \\
    DRAMA & \cite{yuan2024drama}                    & \scriptsize{ISRR'24} & 98.0 & 93.1 & 94.8 & 100  & 80.1 & 85.5 \\ %& Arxiv (2024/08) \\
    VAD-v2 & \cite{chen2024vadv2}                   & \scriptsize{arXiv'24} & 98.1 & 94.8 & 94.3 & 100 & 80.6 & 86.2 \\ %% from tjeir ICLR submission
    PRIX & \cite{wozniak2025prix}                   & \scriptsize{arXiv'25} & 98.1 & 96.3 & 94.1 & 100  & 82.3 & 87.8 \\ %& Arxiv (2025/03)\\
    DiffusionDrive & \cite{liao2025diffusiondrive} & \scriptsize{CVPR'25} & 98.2 & 96.2 & 94.7 & 100  & 82.2 & 88.1 \\ %& Arxiv (2025/03)\\
    DIVER & \cite{song2025breaking}                 & \scriptsize{arXiv'25} & 98.5 & 96.5 & 94.9 & 100  & 82.6 & 88.3 \\ %& Axiv (2025/03)\\
    AutoVLA & \cite{zhou2025autovla}                & \scriptsize{NeurIPS'25} & 98.4 & 95.6 & \textbf{98.0} & 99.9 & 81.9 & 89.1 \\ %& Published \\
    DriveVLA-W0 & \cite{li2025drivevla}             & \scriptsize{ICLR 26'} & 98.7 & \textbf{99.1} & 95.3 & 99.3 & 83.3 & 90.2 \\ %& Arxiv  (2025/10) (ICLR26 submission)\\
    ReCogDrive & \cite{li2025recogdrive}            & \scriptsize{ICLR 26'} & 97.9 & 97.3 & 94.9 & 100  & 87.3 & 90.8 \\ %& Arxiv (2025/06)\\
    Hydra-MDP++   & \cite{li2025hydramdppp}         & \scriptsize{arXiv'25} & 98.6 & 98.6 & 95.1 & 100  & 85.7 & 91.0 \\ %& Arxiv (2025/03) \\
    iPad & \cite{guo2025ipad}                       & \scriptsize{arXiv'25} & 98.6 & 98.3 & 94.9 & 100  & 88.0 & 91.7 \\ %& Arxiv (2025/05)\\
    Centaur & \cite{sima2025centaur}                & \scriptsize{arXiv'25} & 99.5 & 98.9 & \textbf{98.0} & 100  & 85.9 & 92.6 \\ %& Arxiv (2025/03) \\
    DriveSuprim  & \cite{yao2025drivesuprim}         & \scriptsize{arXiv'25} &98.6 & 98.6 & 95.5 & 100 & 91.3 & 93.5 \\%& Arxiv (2025/06) \\
    DrivoR  & \cite{kirby2026driving}         & \scriptsize{CVPR'26} &99.0 & 98.9 & 96.7 & 100 & 90.0 &  93.7 \\%& Arxiv (2025/06) \\
    % \rowcolor{blue!15}
    % \method{}   \tiny{(ViT-S)}  &&& 98.9 & 98.8 & 96.5 & 100 &  89.9 & 93.5        \\
    \rowcolor{blue!15}
    Clover(ours)   &&& \textbf{99.1} & \underline {99.0} & \underline{96.9} & \textbf{100} & \textbf{91.7} &  \textbf{94.5}      \\
    % \rowcolor{blue!15}
    % \method{}    &&& 99.0 & 99.0 & 96.5 & 100 & 90.2 & \bf 93.7      \\
        % Method & & & NC & DAC & TTC & Comf.& EP & \textbf{PDMS} \\
    % RAP-DINO  \tiny{(ViT-H)} & \cite{feng2025rap}                   & \scriptsize{arxiv} & 99.1 & 98.9 & 96.7 & 100 & 90.3 & 93.8 \\ %& Arxiv (2025/10) (ICLR26 submission)\\
    \bottomrule
    \end{tabular}
    % }
    
    \label{tab:benchmark_navsim_v1_small}
\end{table}

\begin{table}[tb]
\centering
\caption{
\textbf{NAVSIM v2 EPDMS on \texttt{navtest}.}
Comparison under the Extended PDMS metric. 
EPDMS$^\ast$ denotes results computed with the original NAVSIM v2 evaluation code before applying the human-behavior filtering fix, while EPDMS denotes results computed with the updated official implementation. 
Higher is better for all columns.
}
\label{tab:navsim_v2}
\resizebox{1\textwidth}{!}{
\begin{tabular}{l|ccccccccc|cc}
    \toprule
    \rowcolor{white}
    Method 
    & NC $\uparrow$ 
    & DAC $\uparrow$ 
    & DDC $\uparrow$ 
    & TL $\uparrow$ 
    & EP $\uparrow$ 
    & TTC $\uparrow$ 
    & LK $\uparrow$ 
    & HC $\uparrow$ 
    & EC $\uparrow$ 
    & \cellcolor{gray!20}EPDMS$^\ast$ $\uparrow$
    & \cellcolor{gray!30}EPDMS $\uparrow$ \\
    \midrule
    Ego-MLP~\cite{li2024ego}
        & 93.1 & 77.9 & 92.7 & 99.6 & 86.0 & 91.5 & 89.4 & \textbf{98.3} & 85.4 
        & \cellcolor{gray!20}64.0 & \cellcolor{gray!30}-- \\
    Transfuser~\cite{chitta2022transfuser} 
        & 96.9 & 89.9 & 97.8 & 99.7 & 87.1 & 95.4 & 92.7 & \textbf{98.3} & 87.2 
        & \cellcolor{gray!20}76.7 & \cellcolor{gray!30}-- \\
    ARTEMIS~\cite{feng2025artemis} 
        & 98.3 & 95.1 & 98.6 & 99.8 & 81.5 & 97.4 & 96.5 & \textbf{98.3} & \textbf{98.3}
        & \cellcolor{gray!20}83.1 & \cellcolor{gray!30}-- \\
    DiffusionDriveV2 ~\cite{zou2025diffusiondrivev2}
        & 97.7 & 96.6 & 99.2 & 99.8 & 88.9 & 97.2 & 96.0 & 97.8 & 91.0 
        & \cellcolor{gray!20}85.5 & \cellcolor{gray!30}87.5 \\
    Hydra-MDP++~\cite{li2025hydramdppp} 
        & 98.4 & 98.0 & 99.4 & 99.8 & 87.5 & 97.7 & 95.3 & \textbf{98.3} & 77.4 
        & \cellcolor{gray!20}85.1 & \cellcolor{gray!30}-- \\
    DriveSuprim~\cite{yao2025drivesuprim} 
        & 97.8 & 97.9 & 99.5 & \textbf{99.9} & 90.6 & 97.1 & 96.6 & \textbf{98.3} & 77.9 
        & \cellcolor{gray!20}86.0 & \cellcolor{gray!30}-- \\
    SparseDriveV2~\cite{sun2026sparsedrivev2scoringneedendtoend}
        & 98.1 & 98.1 & \textbf{99.6} & 99.8 & \textbf{91.1} 
        & 97.3 & \textbf{96.9} & 98.2 & 78.4 
        & \cellcolor{gray!20}86.7 & \cellcolor{gray!30}90.1 \\

    \rowcolor{blue!15}
    Clover(ours)   &\textbf{99.4}&\textbf{99.3}& 99.5 &  99.8 & 86.9 & \textbf{98.9} & 95.2 &  98.2 &75.5&\textbf{87.2}&\textbf{90.4}     \\
    \bottomrule
\end{tabular}
}
\end{table}

\begin{table*}[t]
    \centering
    \caption{
    \textbf{NAVSIM v2 \texttt{navhard-two-stage}.}
    Comparison on the challenging NavHard split using EPDMS. 
    Stage-1 and Stage-2 columns report sub-scores from the two-stage evaluation procedure, and EPDMS denotes the final aggregated score. 
    Higher is better for all columns.
    }
    \label{tab:navhard}
    \begingroup
    \footnotesize
    \setlength{\tabcolsep}{1.15pt}
    \renewcommand{\arraystretch}{0.92}

    \begin{adjustbox}{max width=1\textwidth}
    \begin{tabular}{@{}l@{\hspace{4pt}}*{9}{c}@{\hspace{5pt}}*{9}{c}@{\hspace{5pt}}c@{}}
    \toprule
    \rowcolor{white}
        & \multicolumn{9}{c}{Stage 1}
        & \multicolumn{9}{c}{Stage 2}
        &  \\
    \cmidrule(lr){2-10}
    \cmidrule(lr){11-19}
    Method
        & NC & DAC & DDC & TLC & EP & TTC & LK & HC & EC
        & NC & DAC & DDC & TLC & EP & TTC & LK & HC & EC
        & EPDMS \\
    \midrule

    RAP-DINO~\cite{feng2025rap}
        & 97.1 & 94.4 & 98.8 & 99.8 & 83.9 & 96.9 & 94.7 & 96.4 & 66.2
        & 83.2 & 83.9 & 87.4 & 98.0 & 86.9 & 80.4 & 52.3 & 95.2 & 52.4
        & 39.6 \\

    GTRS-D~\cite{li2025generalized}
        & 98.9 & 96.2 & 99.4 & 99.3 & 72.9 & 98.9 & 95.1 & 96.9 & 39.1
        & 91.2 & 89.4 & 94.4 & 98.8 & 69.5 & 90.0 & 54.3 & 94.0 & 48.7
        & 45.0 \\

    GTRS-A~\cite{li2025generalized}
        & 98.9 & 95.1 & 99.1 & 99.6 & 76.2 & 99.1 & 94.9 & 97.6 & 54.2
        & 88.1 & 88.8 & 89.3 & 98.9 & 98.9 & 85.9 & 53.7 & 96.8 & 56.9
        & 45.4 \\

    DrivoR~\cite{kirby2026driving}
        & 98.8 & 95.1 & 98.9 & 100 & 72.6 & 98.7 & 94.0 & 97.6 & 73.3
        & 90.2 & 88.4 & 91.9 & 98.6 & 70.0 & 88.0 & 50.1 & 98.5 & 76.2
        & 48.3 \\
    \rowcolor{blue!15}
    Clover (ours)
        & 99.4 & 98.0 & 99.9 & 100 & 75.7 & 98.9 & 94.4 & 97.1 & 56.0
        & 89.8 & 88.2 & 93.2 & 99.0 & 74.2 & 87.7 & 52.6 & 97.0 & 57.3
        & 48.3 \\

    \bottomrule
    \end{tabular}
    \end{adjustbox}

    \endgroup
\end{table*}

\begin{table}[t]
\centering
% \small
\setlength{\tabcolsep}{5pt}
\caption{
\textbf{Open-loop planning evaluation on nuScenes.}
We report L2 displacement error and collision rate under the ST-P3 and UniAD protocols. 
Lower is better for all metrics. 
This benchmark is used as supplementary open-loop validation.
}
\label{tab:nuscenes_l2}
\footnotesize
% \resizebox{0.98\textwidth}{!}{
\begin{tabular}{lcccc}
\toprule
\multirow{2}{*}{Method} & \multicolumn{2}{c}{ST-P3 metrics} & \multicolumn{2}{c}{UniAD metrics} \\
\cmidrule(lr){2-3} \cmidrule(lr){4-5}
 & L2 (m) $\downarrow$ & Collision (\%) $\downarrow$ & L2 (m) $\downarrow$ & Collision (\%) $\downarrow$ \\
\midrule
ST-P3~\cite{hu2022st} & 2.11 & 0.71 & -- & -- \\
VAD~\cite{jiang2023vad} & 0.37 & 0.14 & -- & -- \\
UniAD~\cite{hu2023planning} & 0.69 & 0.12 & 1.03 & 0.31 \\
EMMA~\cite{hwang2024emma} & 0.32 & -- & -- & -- \\
OpenEMMA~\cite{xing2025openemma} & 2.81 & -- & -- & -- \\
AutoVLA~\cite{zhou2025autovla} & 0.48 & 0.13 & 0.86 & 0.35 \\
Impromptu VLA~\cite{chi2025impromptu} & 0.33 & 0.13 & 0.67 & 0.38\\
OpenDriveVLA~\cite{zhou2025opendrivevla}& 0.33 & \textbf{0.10} & 0.67 & \textbf{0.30} \\
% Curious-VLA~\cite{chen2026devilnarrowpolicyunleashing}  & 0.31 & 0.10 & \textbf{0.60} & \textbf{0.33} \\

\rowcolor{blue!15}
Clover (ours) &
    \textbf{0.31} & \textbf{0.10} & \textbf{0.65} & \textbf{0.30} \\ 
\bottomrule
\end{tabular}
% }
\end{table}

\subsection{Proposal Quality and Diversity Analysis}
\label{sec:proposal_diversity}

A central goal of CLOVER is to improve the proposal set itself, rather than only rerank a fixed set of candidates.
We therefore analyze the 64 generated proposals across three stages: the DrivoR baseline, Stage 1 pseudo-expert diversity pretraining, and Stage 2 self-distillation refinement.
All statistics are computed on 12,146 common scenes, using true PDMS scores from the evaluator for analysis only.
Following the proposal-diversity protocol of Curious-VLA~\citep{chen2026devilnarrowpolicyunleashing}, we report pairwise ADE/FDE and additional proposal-set quality metrics in Table~\ref{tab:proposal_stage_analysis}.

\begin{figure*}[t]
    \centering
    \includegraphics[width=\textwidth]{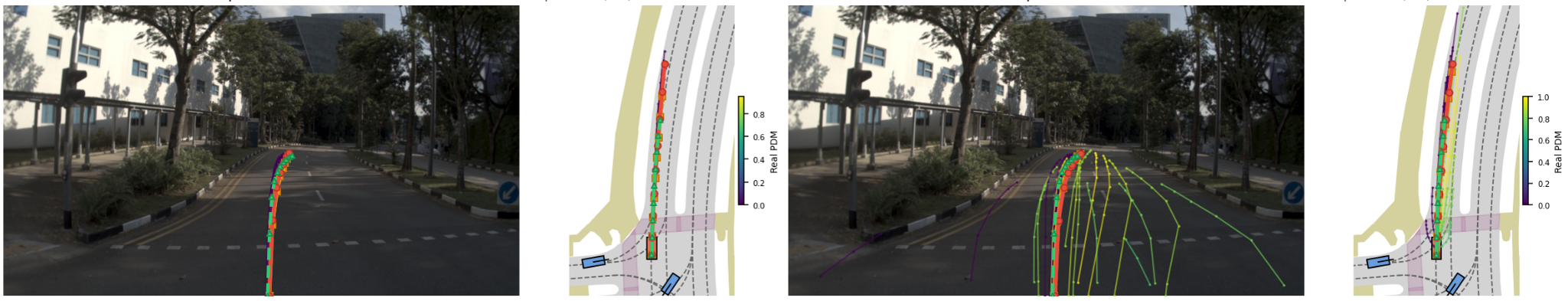}
    \caption{
    \textbf{Qualitative proposal diversity.}
    Compared with the DrivoR baseline, which concentrates candidates around a narrow mode, CLOVER covers a wider set of feasible trajectory branches.
    Many of these diverse candidates remain high-scoring under the evaluator, showing that the additional diversity is not merely caused by low-quality outliers.
    }
    \label{fig:diversity_visualization}
\end{figure*}

Table~\ref{tab:proposal_stage_analysis} shows a clear stage-wise behavior.
Stage 1 greatly expands the proposal space: compared with the baseline, it increases Oracle@64 from 0.9933 to 0.9976, Pairwise ADE from 1.80 to 5.97, Pairwise FDE from 4.47 to 10.70, and Endpoint Area from 5.78 to 46.18.
This confirms that pseudo-expert coverage supervision raises the oracle upper bound, although it also introduces a heavier low-score tail.

Stage 2 then refines this expanded distribution.
It achieves the best scorer-selected PDMS, improving from 0.9369 for the baseline and 0.9413 for Stage 1 to 0.9448.
It also obtains the highest mean proposal score, 0.8277, and reduces the number of low-score proposals with PDMS$<0.50$ to 6.83.
At the same time, it preserves substantially more diversity than the baseline, with Pairwise ADE 5.20 versus 1.80 and qualified cluster count 8.71 versus 6.02.
These results support the intended division of labor: Stage 1 expands proposal support, while Stage 2 converts the expanded space into a higher-quality and better-ranked proposal distribution.
Additional qualitative comparisons and scorer-only diagnostic experiments are provided in Appendices~\ref{app:proposal_diversity_qualitative} and~\ref{app:scorer_diagnostic}.

\begin{table*}[t]
\centering
\small
\caption{
\textbf{Stage-wise proposal quality and diversity analysis.}
Metrics are computed over 64 generated proposals on 12,146 common scenes. 
Stage 1 greatly expands the proposal space and achieves the highest oracle upper bound, while Stage 2 refines the expanded distribution, obtaining the best selected score and proposal-set quality while preserving much higher diversity than the baseline. 
Higher is better except for Gap and Count(PDMS$<0.50$).
}
\label{tab:proposal_stage_analysis}
\begin{tabular}{lccc}
\toprule
Metric & Baseline & Stage 1 & Stage 2 \\
\midrule
Selected PDMS $\uparrow$ & 0.9369 & 0.9413 & \textbf{0.9448} \\
Oracle@64 PDMS $\uparrow$ & 0.9933 & \textbf{0.9976} & 0.9939 \\
Gap(Oracle - Selected) $\downarrow$ & 0.0564 & 0.0563 & \textbf{0.0495} \\
Mean PDMS over 64 $\uparrow$ & 0.7972 & 0.7374 & \textbf{0.8277} \\
Std PDMS over 64 & 0.2152 & \textbf{0.3013} & 0.2235 \\
Count(PDMS$>0.95$) $\uparrow$ & 30.94 & 24.78 & \textbf{31.97} \\
Count(PDMS$<0.50$) $\downarrow$ & 9.05 & 12.21 & \textbf{6.83} \\
\midrule
Pairwise ADE $\uparrow$ & 1.80 & \textbf{5.97} & 5.20 \\
Pairwise FDE $\uparrow$ & 4.47 & \textbf{10.70} & 8.25 \\
Endpoint Area $\uparrow$ & 5.78 & \textbf{46.18} & 25.56 \\
Trajectory Effective Rank $\uparrow$ & 1.14 & 1.24 & \textbf{1.27} \\
Qualified Cluster Count@2m $\uparrow$ & 6.02 & \textbf{10.91} & 8.71 \\
Top-6 Real Pairwise ADE $\uparrow$ & 0.91 & \textbf{2.11} & 1.75 \\
Top-6 Real Pairwise FDE $\uparrow$ & 2.20 & \textbf{3.55} & 2.59 \\
\bottomrule
\end{tabular}
\end{table*}

\subsection{Ablation Studies}
\label{sec:ablation}

We ablate the main components of CLOVER on NAVSIM v1 PDMS and NAVSIM v2 EPDMS.
The ablations examine four design choices: pseudo-expert proposal coverage, alternating closed-loop refinement, vector-Pareto teacher-set construction, and anchor-assisted reranking.

\textbf{Main components.}
Table~\ref{tab:ablation_main_components} shows that the two stages are complementary.
The single-expert baseline obtains 93.7 PDMS.
Adding pseudo-expert diversity supervision improves the score to 94.1, while applying closed-loop refinement without the diversity stage yields only 93.8.
The full model reaches 94.5, suggesting that scorer-guided refinement is most effective when Stage 1 has already expanded the proposal support.

\begin{table}[t]
\centering
\small
\caption{
\textbf{Main component ablation on NAVSIM v1.}
We ablate pseudo-expert diversity supervision and Stage-2 closed-loop refinement. 
Diversity supervision provides a stronger proposal set, while closed-loop refinement is most effective when applied on top of the expanded proposal distribution.
}
\label{tab:ablation_main_components}
\begin{tabular}{lccc}
\toprule
Setting & Diversity Sup. & Stage-2 Refinement & PDMS $\uparrow$ \\
\midrule
Single-expert supervision & -- & -- & 93.7 \\
Diversity only & \checkmark & -- & 94.1 \\
Closed-loop only & -- & \checkmark & 93.8 \\
CLOVER full & \checkmark & \checkmark & \textbf{94.5} \\
\bottomrule
\end{tabular}
\end{table}

\textbf{Stage-2 refinement design.}
Table~\ref{tab:ablation_stage2} studies the training strategy for closed-loop refinement.
A non-iterative generator/scorer update collapses severely, indicating that directly updating the generator against a changing imperfect scorer is unstable.
Self-distillation alone remains stable but saturates at 93.8--94.0 PDMS, depending on the number of distilled trajectories.
The full alternating procedure achieves 94.5, supporting the need to repeatedly refit the scorer to evaluator-provided sub-scores before using it to guide generator refinement.

\begin{table}[t]
\centering
\small
\caption{
\textbf{Ablation of Stage-2 refinement.}
Non-iterative joint updating causes severe distribution drift. 
Self-distillation without scorer fitting remains stable but underperforms the full alternating scorer-generator refinement.
}
\label{tab:ablation_stage2}
\begin{tabular}{lc}
\toprule
Stage-2 variant & PDMS $\uparrow$ \\
\midrule
Non-iterative generator/scorer update & $<10$ \\
Self-distillation only, 64 trajectories & 94.0 \\
Self-distillation only, 32 trajectories & 93.8 \\
Self-distillation only, 16 trajectories & 93.8 \\
Self-distillation only, 8 trajectories & 93.8 \\
Full alternating closed-loop refinement & \textbf{94.5} \\
\bottomrule
\end{tabular}
\end{table}

\textbf{Teacher-set construction.}
Table~\ref{tab:ablation_teacher_set} compares different target sets for Stage-2 generator refinement.
Using scalar real-PDMS top-$k$ targets gives 93.9 PDMS and relatively low proposal diversity, suggesting that scalar top-$k$ supervision can over-concentrate the student around a narrow set of high-score modes.
Adding distance suppression improves both PDMS and diversity, but still underperforms vector-Pareto guidance.
Our vector-Pareto targets achieve the best PDMS and the highest high-quality proposal diversity, indicating that preserving multi-objective trade-offs among safety, progress, and comfort is beneficial for proposal refinement.

\begin{table}[t]
\centering
\small
\caption{
\textbf{Ablation of teacher-set construction in Stage 2.}
Scalar real-PDMS top-$k$ targets tend to concentrate proposals around a narrow high-score mode. 
Distance suppression partially restores diversity, while vector-Pareto guidance achieves the best PDMS and is expected to preserve more diverse high-quality proposal modes.
Q-Cluster denotes Qualified Cluster Count@2m computed among proposals with PDMS$\geq0.8$.
Top-6 ADE/FDE are computed among the six highest real-PDMS proposals.
}
\label{tab:ablation_teacher_set}
\begin{tabular}{lcccc}
\toprule
Teacher target construction 
& PDMS $\uparrow$ 
& Q-Cluster@2m $\uparrow$ 
& Top-6 ADE $\uparrow$ 
& Top-6 FDE $\uparrow$ \\
\midrule
Real-PDMS top-$k$ & 93.9 & 6.45 & 1.04 & 2.31 \\
Real-PDMS top-$k$ + distance suppression & 94.3 & 7.49 & 1.66 & 2.47 \\
Vector-Pareto targets (ours) & \textbf{94.5} & 8.71 & 1.75 & 2.59 \\
\bottomrule
\end{tabular}
\end{table}

\textbf{Anchor-assisted reranking.}
Table~\ref{tab:ablation_anchor} evaluates the optional anchor-assisted soft reranking used for NAVSIM v2 EPDMS.
The single-mode anchor alone is weaker than CLOVER, and CLOVER without anchor already obtains 89.3 EPDMS.
Using the anchor as a soft reference improves EPDMS to 90.4 and substantially improves EC, supporting its role in reducing frame-to-frame selection jitter among near-tied high-quality candidates.
The full anchor weight sweep is provided in Appendix~\ref{app:anchor_soft}.

\textbf{Number of proposals.}
We also ablate the number of generated trajectories.
Using $K=32$, $64$, and $96$ proposals yields 93.9, 94.5, and 94.4 PDMS, respectively.
Thus, increasing the proposal set from 32 to 64 improves candidate coverage, while further increasing to 96 brings no additional gain and may slightly increase ranking difficulty; we use $K=64$ by default.

\begin{table}[t]
\centering
\small
\caption{
\textbf{Anchor-assisted soft reranking ablation on NAVSIM v2 EPDMS.}
The anchor model alone is weaker than CLOVER, but using it as a soft reference improves EPDMS by reducing diversity-induced selection jitter.
}
\label{tab:ablation_anchor}
\begin{tabular}{lcc}
\toprule
Inference setting  & EC $\uparrow$ & EPDMS $\uparrow$ \\
\midrule
Single-mode anchor only  & 77.6 & 88.7 \\
CLOVER without anchor  & 35.0 & 89.3 \\
CLOVER + anchor-assisted soft reranking  & 75.5 & \textbf{90.4} \\
\bottomrule
\end{tabular}
\end{table}

\section{Limitations and Future Work}
\label{sec:limitations}

CLOVER currently performs scoring and ranking mainly at the per-scene trajectory level. 
While effective for PDMS-style evaluation, temporally aggregated EPDMS terms such as extended comfort depend on cross-frame consistency, which is only mitigated by our optional anchor-assisted reranking. 
Future work could integrate history-aware or sequence-level scorers to model temporal consistency directly during ranking and refinement.

\section{Conclusion}
\label{sec:conclusion}

We presented \textbf{CLOVER}, a closed-loop value estimation and ranking framework for proposal-selection based end-to-end autonomous driving planning.
CLOVER addresses the training--evaluation mismatch by first expanding single-trajectory imitation into evaluator-filtered set-level proposal coverage, and then refining the generator through conservative closed-loop self-distillation.
A trajectory-level scorer is fitted to true evaluator sub-scores on generated proposals, while teacher-selected top-$k$ and vector-Pareto targets guide the generator with stability regularization.
We further analyzed when an imperfect scorer can reliably improve proposal generation, showing that conservative refinement can increase high-quality proposal support when scorer-selected targets are statistically enriched under the true evaluator.
Experiments on NAVSIM v1, NAVSIM v2 EPDMS, NavHard, and supplementary nuScenes evaluation demonstrate strong planning performance, while proposal-quality analyses show that CLOVER improves not only final top-1 selection but also the underlying candidate distribution.

% \section*{References}

\bibliography{example_paper}
\bibliographystyle{plainnat}

% References follow the acknowledgments in the camera-ready paper. Use unnumbered first-level heading for
% the references. Any choice of citation style is acceptable as long as you are
% consistent. It is permissible to reduce the font size to \verb+small+ (9 point)
% when listing the references.
% Note that the Reference section does not count towards the page limit.
% \medskip

% {
% \small

% [1] Alexander, J.A.\ \& Mozer, M.C.\ (1995) Template-based algorithms for
% connectionist rule extraction. In G.\ Tesauro, D.S.\ Touretzky and T.K.\ Leen
% (eds.), {\it Advances in Neural Information Processing Systems 7},
% pp.\ 609--616. Cambridge, MA: MIT Press.

% [2] Bower, J.M.\ \& Beeman, D.\ (1995) {\it The Book of GENESIS: Exploring
%   Realistic Neural Models with the GEneral NEural SImulation System.}  New York:
% TELOS/Springer--Verlag.

% [3] Hasselmo, M.E., Schnell, E.\ \& Barkai, E.\ (1995) Dynamics of learning and
% recall at excitatory recurrent synapses and cholinergic modulation in rat
% hippocampal region CA3. {\it Journal of Neuroscience} {\bf 15}(7):5249-5262.
% }

%%%%%%%%%%%%%%%%%%%%%%%%%%%%%%%%%%%%%%%%%%%%%%%%%%%%%%%%%%%%

\appendix

% \section{Technical appendices and supplementary material}
% Technical appendices with additional results, figures, graphs, and proofs may be submitted with the paper submission before the full submission deadline (see above). You can upload a ZIP file for videos or code, but do not upload a separate PDF file for the appendix. There is no page limit for the technical appendices. 

% Note: Think of the appendix as ``optional reading'' for reviewers. The paper must be able to stand alone without the appendix; for example, adding critical experiments that support the main claims to an appendix is inappropriate. 

\newpage

\section{Pseudo-Expert Trajectory Generation}
\label{app:pseudo_expert_generation}

CLOVER uses pseudo-expert trajectories to provide set-level multi-modal supervision beyond the single logged trajectory. 
The pseudo-expert pipeline is used only during training. 
It constructs a diverse trajectory pool using privileged metric-cache information, scores and filters the candidates with the NAVSIM evaluator, and converts the resulting pool into compact per-scene supervision targets during Stage-1 training. 
The overall data flow is:
\[
\begin{aligned}
\text{metric cache}
&\rightarrow \text{candidate family generation} \\
&\rightarrow \text{drivable-area / occupancy pre-check} \\
&\rightarrow \text{PDM scoring} \\
&\rightarrow \text{coverage-aware selection} \\
&\rightarrow \text{post-processing} \\
&\rightarrow \text{training-time pseudo-expert sampling}
\end{aligned}
\]

\paragraph{Training-time privileged information.}
The pseudo-expert generator uses information that is available in the training metric cache but not available to the deployed planner. 
This includes the current ego state, route centerline, drivable-area map, future obstacle occupancy, and the NAVSIM PDM evaluator. 
Therefore, pseudo-experts should be understood as privileged training supervision rather than an online planner. 
At inference time, CLOVER only uses the learned generator and scorer.

\paragraph{Trajectory parameterization.}
Each candidate trajectory is generated in a centerline-aligned coordinate system. 
Let $d_0$ be the current ego progress along the route centerline and let $\mathrm{lat}_0$ be the current lateral offset. 
A candidate is defined by a longitudinal progress curve $d_t$ and a lateral offset curve $\mathrm{lat}_t$, and then mapped back to Cartesian coordinates:
\begin{equation}
    d_t = d_0 + \sum_{k \leq t} v_k \Delta t ,
\end{equation}
\begin{equation}
    x_t = x_{\mathrm{cl}}(d_t) - \mathrm{lat}_t \sin h_{\mathrm{cl}}(d_t),
    \quad
    y_t = y_{\mathrm{cl}}(d_t) + \mathrm{lat}_t \cos h_{\mathrm{cl}}(d_t),
\end{equation}
\begin{equation}
    \theta_t = h_{\mathrm{cl}}(d_t),
\end{equation}
where $(x_{\mathrm{cl}},y_{\mathrm{cl}},h_{\mathrm{cl}})$ denotes the route centerline pose. 
For lateral transitions, we use a quintic smooth step:
\begin{equation}
    s(r) = 6r^5 - 15r^4 + 10r^3,
\end{equation}
which has zero first- and second-order derivatives at both endpoints. 
This produces smoother lateral shifts than linear interpolation and enables different comfort profiles even when the initial and final lateral offsets are similar.

\paragraph{Candidate action families.}
For each scene, we generate a structured pool of candidates from six interpretable action families, summarized in Table~\ref{tab:pseudo_expert_families}. 
These families are designed to cover both longitudinal and lateral variations that affect PDM sub-scores, including progress, time-to-collision, comfort, and drivable-area compliance. 
The goal is not to keep only high-score candidates, but to expose the model to a broad range of feasible, near-feasible, and informative low-score behaviors.

\begin{table}[H]
\centering
\small
\caption{
\textbf{Pseudo-expert candidate families.}
Each scene first generates a structured pool of candidate trajectories from interpretable longitudinal and lateral action families.
}
\label{tab:pseudo_expert_families}
\begin{tabular}{lcl}
\toprule
Family & Count & Purpose \\
\midrule
Constant-speed lateral transitions & 200 & Cover smooth and aggressive lateral shifts under different speeds \\
Off-road boundary cases & 12 & Populate low-score drivable-area violation regions \\
Acceleration profiles & 18 & Vary longitudinal progress and speed profiles \\
Stop-go behaviors & 9 & Cover safe but low-progress behaviors \\
Approach-brake behaviors & 10 & Create TTC / comfort trade-offs \\
Overshoot lateral motions & 12 & Add aggressive lateral dynamics and comfort variation \\
\midrule
Total & 261 & Structured candidate pool per scene \\
\bottomrule
\end{tabular}
\end{table}

\paragraph{Pre-check before PDM scoring.}
Running the full PDM evaluator on every raw candidate is expensive. 
We therefore first perform lightweight geometric pre-checks. 
For drivable-area compliance, we compute ego-vehicle corners at each future pose and query whether they lie within valid drivable polygons. 
For collision pre-checking, we compare the future ego footprint with future obstacle occupancy from the metric cache. 
Each trajectory is assigned one of three coarse labels: feasible, near-feasible, or infeasible. 
We prioritize feasible and near-feasible candidates for full PDM scoring, while retaining a small subset of infeasible candidates to preserve score coverage. 
In our implementation, at most 180 candidates per scene are sent to the full evaluator.

\paragraph{PDM scoring and sub-score recording.}
Each selected candidate is evaluated by the NAVSIM PDM evaluator. 
Besides the final scalar PDM score, we also store its sub-scores, including no-at-fault collisions, drivable-area compliance, ego progress, time-to-collision, comfort, and driving-direction compliance. 
These sub-scores are used both for scorer supervision and for coverage-aware pseudo-expert selection.

\paragraph{Coverage-aware selection.}
A naive top-score selection would often keep many similar high-score trajectories, causing the pseudo-expert set to collapse back to a narrow mode. 
Instead, we select a compact per-scene set by covering different PDM sub-score patterns and score ranges. 
For each scored candidate, we define a coverage key:
\begin{equation}
    \mathrm{key}
    =
    \left(
    \mathrm{DAC},
    \mathrm{Collision},
    \mathrm{TTC},
    \mathrm{Comfort},
    \mathrm{ProgressBin}
    \right),
\end{equation}
where binary or discrete PDM components are rounded to their categorical values, and ego progress is discretized into progress bins. 
We then greedily select candidates by minimizing:
\begin{equation}
    \mathrm{cost}_i
    =
    10 \cdot \mathrm{cov\_count}(\mathrm{key}_i)
    +
    \mathrm{bin\_count}(\mathrm{score\_bin}_i),
\end{equation}
where $\mathrm{cov\_count}$ counts how many already selected candidates share the same sub-score key, and $\mathrm{bin\_count}$ counts how many candidates have been selected from the same total-score bin. 
This encourages coverage over both sub-score combinations and scalar score ranges. 
We keep up to 50 trajectories per scene in the pseudo-expert dataset before post-processing.

\paragraph{Post-processing and boundary interpolation.}
The generated pseudo-expert dataset is further cleaned before training. 
We remove candidates with wrong driving direction, i.e., trajectories whose driving-direction compliance is less than 1.0. 
We then identify score-boundary regions where small changes in lateral offset cause a large PDM drop. 
Specifically, within each family and speed group, adjacent lateral targets are compared; if the PDM score drop exceeds a threshold, we interpolate an additional candidate between the high-score and low-score sides and evaluate it again. 
These boundary-interpolation samples provide fine-grained supervision near metric discontinuities, where small geometric changes can lead to large score differences.

\paragraph{Training-time pseudo-expert sampling.}
The stored pseudo-expert pool is not used directly in full during training. 
For each scene, we first keep trajectories with PDM score above a threshold, then sort them by score and apply greedy farthest-point sampling in trajectory space. 
This produces at most $M=8$ pseudo-experts per scene. 
If the logged human trajectory is not sufficiently covered by the selected pseudo-expert set, it is added to preserve the human driving prior. 
The selected trajectories are returned to the model as:
\[
    \texttt{pseudo\_experts} \in \mathbb{R}^{M \times T \times 3},
    \quad
    \texttt{pseudo\_expert\_mask} \in \{0,1\}^{M}.
\]
If no valid pseudo-expert is available for a scene, the system falls back to the logged human trajectory. 
This ensures that the training schema remains consistent while retaining standard imitation supervision as a safe fallback.

\paragraph{Key hyperparameters.}
Table~\ref{tab:pseudo_expert_key_hparams} summarizes the main hyperparameters used in pseudo-expert generation and training-time sampling.

\begin{table}[t]
\centering
\small
\caption{
\textbf{Key hyperparameters for pseudo-expert generation and sampling.}
The candidate pool is generated from structured action families, scored by the evaluator, filtered and selected for coverage, and finally sampled into a compact training-time pseudo-expert set.
}
\label{tab:pseudo_expert_key_hparams}
\begin{tabular}{ll}
\toprule
Hyperparameter & Value \\
\midrule
Number of future poses & 8 \\
Sampling interval & 0.5 s \\
Speed candidates & $(0,2,4,6,8,10,12,15)$ m/s \\
Regular lateral offsets & $(-3.5,-2.0,-1.0,-0.5,0,0.5,1.0,2.0,3.5)$ m \\
Off-road lateral offsets & $(-7.0,-5.5,5.5,7.0)$ m \\
Transition portions & $(0.35,0.6,1.0)$ \\
Acceleration rates & $(-2.0,-1.0,-0.5,0.5,1.0,2.0)$ m/s$^2$ \\
Raw candidates per scene & 261 \\
Maximum PDM-scored candidates per scene & 180 \\
Retained pseudo-expert pool per scene & 50 \\
Score-drop threshold for boundary interpolation & 0.25 \\
Maximum score boundaries per scene & 3 \\
Boundary interpolation samples per boundary & 1 \\
Training-time score threshold & 0.8 \\
Training-time pseudo-expert top-$K$ & 8 \\
\bottomrule
\end{tabular}
\end{table}

\begin{figure*}[p]
    \centering
    \includegraphics[height=0.85\textheight,keepaspectratio]{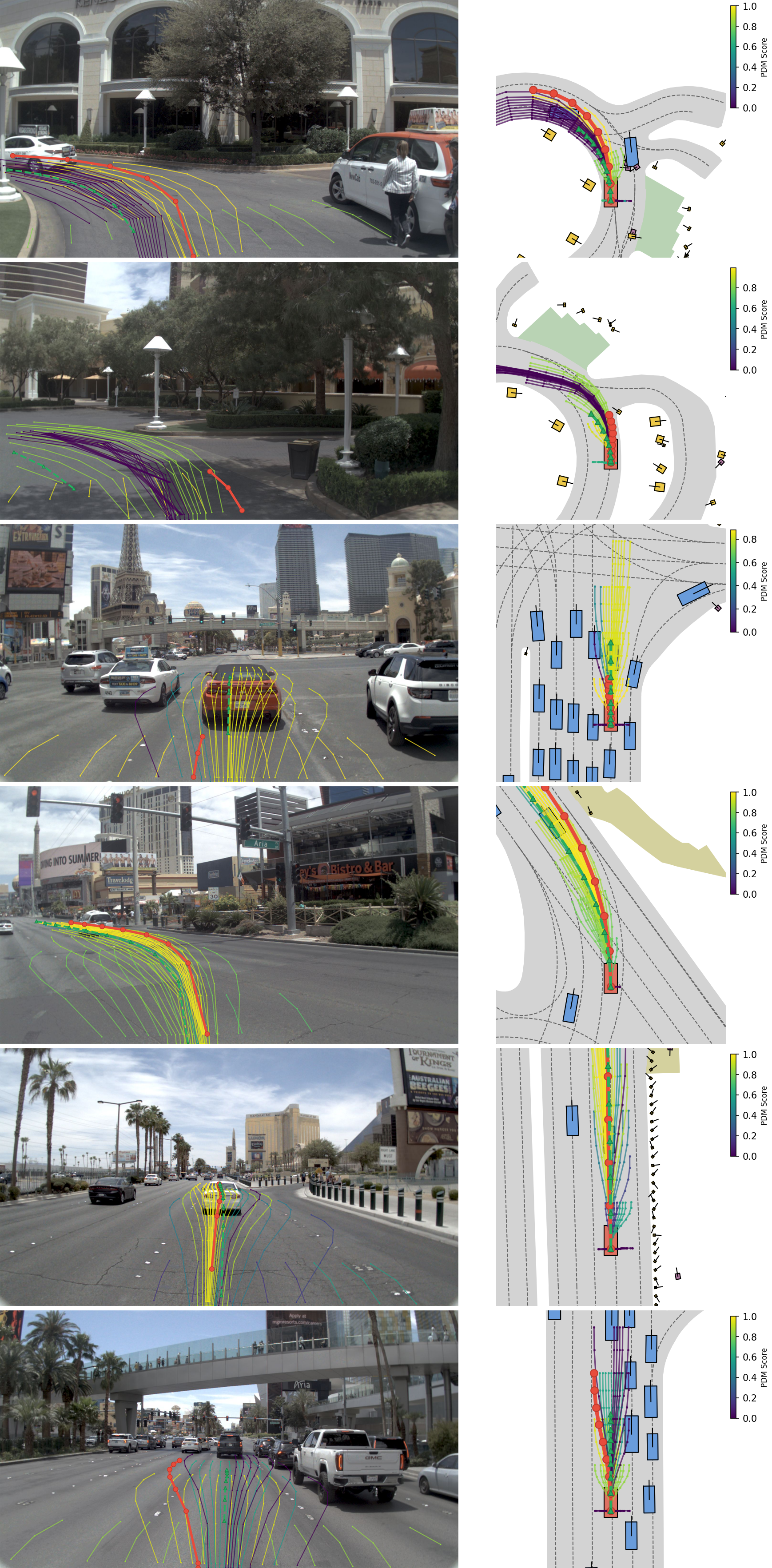}
    \caption{
    \textbf{Visualization of pseudo-expert trajectory candidates.}
    We show representative pseudo-expert trajectory pools in camera view and bird's-eye view. 
    Candidate trajectories are colored by their true PDM scores. 
    The pseudo-expert generator creates diverse trajectories covering different lateral offsets, progress regimes, and boundary cases. 
    High-score candidates provide feasible multi-modal supervision, while lower-score and boundary candidates help expose the generator and scorer to drivable-area, collision, progress, and comfort variations.
    These pseudo-experts are used only during Stage-1 training and are not required at inference time.
    }
    \label{fig:pseudo_expert_visualization}
\end{figure*}

\paragraph{Qualitative examples.}
Figure~\ref{fig:pseudo_expert_visualization} visualizes pseudo-expert trajectory pools on representative scenes. 
The candidates cover a wide range of lateral offsets, progress profiles, and safety/comfort trade-offs. 
The visualization also illustrates why pseudo-expert supervision is more informative than local perturbation around the logged trajectory: the generated pool contains both high-score feasible alternatives and lower-score boundary cases that expose the model to metric discontinuities.

\section{Trajectory-quality Score in NAVSIM (PDMS v1 and EPDMS v2)}
\label{app:offline_score}

We use the planning-oriented NAVSIM benchmark and adopt the official Predictive Driver Model Score (PDMS) from NAVSIM v1 as our primary offline trajectory-quality score $s(\cdot,\cdot)$; higher is better. PDMS is a pseudo closed-loop metric that holistically assesses safety, comfort, and progress via multiplicative penalties and a weighted average:
\begin{equation}
\label{eq:PDMS}
\mathrm{PDMS} \;=\; \mathrm{NC} \times \mathrm{DAC} \times
\left(\frac{5\cdot \mathrm{EP} + 5\cdot \mathrm{TTC} + 2\cdot \mathrm{C}}{12}\right),
\end{equation}
where $\mathrm{NC}$ denotes no at-fault collisions, $\mathrm{DAC}$ drivable-area compliance, $\mathrm{EP}$ ego progress, $\mathrm{TTC}$ time-to-collision within bound, and $\mathrm{C}$ comfort.

\paragraph{NAVSIM v2: Extended PDMS (EPDMS).}
NAVSIM v2 extends PDMS to improve coverage and fairness of open-loop planning evaluation. Compared to NAVSIM v1, EPDMS introduces additional weighted subscores (lane keeping and extended comfort variants), additional multiplier penalties (driving direction compliance and traffic light compliance), and a false-positive penalty filtering scheme.

\begin{table}[t]
\centering
\small
\caption{EPDMS composition in NAVSIM v2 (new metrics relative to v1 are highlighted).}
\label{tab:ePDMS_metrics}
\begin{tabular}{l c c}
\toprule
Metric & Weight & Range \\
\midrule
No at-fault Collisions (NC) & multiplier & $\{0, \tfrac{1}{2}, 1\}$ \\
Drivable Area Compliance (DAC) & multiplier & $\{0, 1\}$ \\
\textbf{Driving Direction Compliance (DDC)} & multiplier & $\{0, \tfrac{1}{2}, 1\}$ \\
\textbf{Traffic Light Compliance (TLC)} & multiplier & $\{0, 1\}$ \\
Ego Progress (EP) & 5 & $[0,1]$ \\
Time to Collision (TTC) within bound & 5 & $\{0,1\}$ \\
\textbf{Lane Keeping (LK)} & 2 & $\{0,1\}$ \\
\textbf{History Comfort (HC)} & 2 & $\{0,1\}$ \\
\textbf{Extended Comfort (EC)} & 2 & $\{0,1\}$ \\
\bottomrule
\end{tabular}
\end{table}

\paragraph{False-positive penalty filtering.}
To reduce false-positive penalties, NAVSIM v2 disables a penalty when the human agent is also responsible for the corresponding violation. Formally, for a metric $m$, define
\begin{equation}
\label{eq:ePDMS_filter}
\mathrm{filter}_m(\mathrm{agent},\mathrm{human}) =
\begin{cases}
1.0, & \text{if } m(\mathrm{human}) = 0,\\
m(\mathrm{agent}), & \text{otherwise.}
\end{cases}
\end{equation}
Intuitively, if the human baseline also triggers the violation, the metric is neutralized (set to $1.0$) rather than penalizing the planner.

\paragraph{EPDMS definition.}
With the above filtering, EPDMS is defined as
\begin{align}
\label{eq:ePDMS}
\mathrm{EPDMS} \;=\;&
\left(\prod_{m\in\{\mathrm{NC},\mathrm{DAC},\mathrm{DDC},\mathrm{TLC}\}} \mathrm{filter}_m(\mathrm{agent},\mathrm{human})\right)
\\
&\cdot
\left(
\frac{\sum_{m\in\{\mathrm{TTC},\mathrm{EP},\mathrm{HC},\mathrm{LK},\mathrm{EC}\}} w_m \cdot \mathrm{filter}_m(\mathrm{agent},\mathrm{human})}
{\sum_{m\in\{\mathrm{TTC},\mathrm{EP},\mathrm{HC},\mathrm{LK},\mathrm{EC}\}} w_m}
\right),
\end{align}
where $w_m$ are the weights listed in Table~\ref{tab:ePDMS_metrics}.

\paragraph{Pseudo closed-loop aggregation in NAVSIM v2.}
NAVSIM v1 computes metrics after a 4-second non-reactive simulation rollout (background actors follow recorded futures; ego follows the planned trajectory via a controller). NAVSIM v2 uses a two-stage aggregation to better approximate closed-loop behavior while remaining open-loop:
(i) a first-stage score is computed on an initial 4-second scene;
(ii) multiple follow-up scenes (precomputed rollouts starting from the same initial scene but with different end states) are also scored, and then aggregated with weights given by a Gaussian kernel based on how close each follow-up scene's start state is to the submitted planner's first-stage end state.
Finally, the first-stage score and the aggregated second-stage score are multiplied to obtain the final aggregated EPDMS score.

\section{Anchor-assisted Soft Reranking}
\label{app:anchor_soft}

This appendix describes the optional anchor-assisted soft reranking used in our NAVSIM v2 EPDMS submission.
The motivation is that a strongly diverse proposal set can improve oracle coverage, but may also cause the selected trajectory to vary across adjacent evaluation frames.
This can reduce extended comfort in EPDMS.
We therefore introduce a lightweight anchor reference to encourage temporally smoother selection, without using future information or evaluator scores at inference time.

\paragraph{Anchor model.}
The anchor trajectory is produced by a single-mode planner with the same observation inputs and the same base architecture as CLOVER.
The anchor model is trained with diversity-related settings disabled and standard single-trajectory imitation retained.
It provides a stable reference trajectory, but is not used as an evaluator and does not filter candidates.

\paragraph{Soft reranking formula.}
For each candidate trajectory $\tau_k$, let
\[
\mathbf{p}_{k}(t)=(x_{k}(t),y_{k}(t),\psi_{k}(t)),
\quad
t=1,\ldots,T,
\]
and let the anchor trajectory be
\[
\mathbf{a}(t)=(x_a(t),y_a(t),\psi_a(t)).
\]
We define the full-trajectory position RMS:
\begin{equation}
    \mathrm{XYRMS}_k
    =
    \sqrt{
    \frac{1}{T}
    \sum_{t=1}^{T}
    \left[
    (x_k(t)-x_a(t))^2
    +
    (y_k(t)-y_a(t))^2
    \right]
    },
\end{equation}
and the full-trajectory heading RMS:
\begin{equation}
    \mathrm{HeadRMS}_k
    =
    \sqrt{
    \frac{1}{T}
    \sum_{t=1}^{T}
    \operatorname{wrap}
    \left(
    \psi_k(t)-\psi_a(t)
    \right)^2
    }.
\end{equation}
The anchor penalty is:
\begin{equation}
    \mathrm{Penalty}_k
    =
    \lambda_{xy}
    \frac{\mathrm{XYRMS}_k}{S_{\mathrm{pos}}}
    +
    \lambda_{\psi}
    \frac{\mathrm{HeadRMS}_k}{S_{\psi}},
\end{equation}
where $S_{\mathrm{pos}}=5.0$ and $S_{\psi}=0.35$.
Given the learned scorer score $s_k$, the final soft reranking score is:
\begin{equation}
    Q_k
    =
    \lambda_s s_k
    -
    \mathrm{Penalty}_k.
\end{equation}
The selected candidate is:
\begin{equation}
    k^\ast = \arg\max_k Q_k.
\end{equation}
This is a soft scheme: no hard threshold is used to discard candidates.

\paragraph{Parameter search.}
We sweep the scorer weight $\lambda_s$, position-anchor weight $\lambda_{xy}$, and heading-anchor weight $\lambda_{\psi}$ on the validation/evaluation split.
The main reported result uses:
\[
    \lambda_s=2.00,\quad
    \lambda_{xy}=0.20,\quad
    \lambda_{\psi}=0.50.
\]
This configuration achieves the best score in our search, with a final score of $0.90356$.

\begin{table}[t]
\centering
% \scriptsize
\setlength{\tabcolsep}{4pt}
\renewcommand{\arraystretch}{0.95}
\caption{
\textbf{Coarse sweep for anchor-assisted soft reranking.}
We report representative grid-search results over scorer weight $\lambda_s$, position-anchor weight $\lambda_{xy}$, and heading-anchor weight $\lambda_{\psi}$.
Scores are normalized EPDMS values.
The no-anchor baseline obtains $0.8930$ EPDMS, while a broad range of anchor configurations consistently improves performance.
The main result uses $\lambda_s=2.00$, $\lambda_{xy}=0.20$, and $\lambda_{\psi}=0.50$.
}
\label{tab:anchor_soft_search}
\begin{tabular}{ccccccc}
\toprule
$\lambda_s$ & $\lambda_{xy}$ & $\lambda_{\psi}$ & Score & EP & EC & HC \\
\midrule
--   & --   & --   & 0.89322 & \textbf{0.92408}    & 0.3500 & 0.96459     \\
\midrule
2.00 & 0.20 & 0.50 & \textbf{0.90356} & 0.8689 & 0.7550 & 0.9823 \\
\midrule
1.75 & 0.20 & 0.75 & 0.90213 & 0.8613 & 0.7641 & 0.9825 \\
2.00 & 0.20 & 0.75 & 0.90292 & 0.8652 & 0.7594 & 0.9825 \\
2.25 & 0.20 & 0.75 & 0.90222 & 0.8689 & 0.7509 & 0.9823 \\
\midrule
1.75 & 0.20 & 1.00 & 0.90212 & 0.8596 & 0.7667 & 0.9825 \\

2.00 & 0.20 & 1.00 & 0.90212 & 0.8630 & 0.7605 & 0.9825 \\
2.25 & 0.20 & 1.00 & 0.90168 & 0.8663 & 0.7539 & 0.9824 \\
\midrule

2.00 & 0.25 & 0.50 & 0.90285 & 0.8633 & 0.7610 & 0.9825 \\

2.00 & 0.25 & 0.75 & 0.90194 & 0.8601 & 0.7640 & 0.9825 \\

\midrule
1.75 & 0.35 & 0.75 & 0.90086 & 0.8502 & 0.7698 & 0.9828 \\
2.00 & 0.35 & 0.75 & 0.90177 & 0.8534 & 0.7679 & 0.9826 \\
2.25 & 0.35 & 0.75 & 0.90216 & 0.8564 & 0.7660 & 0.9825 \\
\midrule
1.75 & 0.35 & 1.00 & 0.90021 & 0.8492 & \textbf{0.7718} & \textbf{0.9828} \\
2.00 & 0.35 & 1.00 & 0.90118 & 0.8521 & 0.7699 & 0.9826 \\
2.25 & 0.35 & 1.00 & 0.90180 & 0.8549 & 0.7679 & 0.9825 \\
\bottomrule
\end{tabular}
\end{table}

The search shows a consistent trade-off between final score and extended comfort.
A light position anchor, around $\lambda_{xy}=0.20$, preserves progress and achieves the best total score, while stronger heading or position anchoring can improve comfort at the cost of progress and final score.

\section{Theoretical Analysis}
\label{app:theory}

This appendix provides full proofs for the theoretical results in Section~\ref{sec:theory}.
All scores are normalized to $[0,1]$.
For simplicity, we write expectations over $o\sim\mathcal{D}$ implicitly when no ambiguity arises.

\subsection{Notation}
\label{app:theory_notation}

Let $o\sim\mathcal{D}$ be a logged driving scene and let $\tau\in\mathcal{X}_o$ be a candidate ego trajectory.
The true non-differentiable trajectory-quality score is
\begin{equation}
    R^*(o,\tau)\in[0,1].
\end{equation}
The learned scalar scorer at iteration $t$ is
\begin{equation}
    s_t(o,\tau)
    =
    \Gamma_{\mathrm{PDMS}}\!\left(S_t(o,\tau)\right)
    \in[0,1],
\end{equation}
where $S_t(o,\tau)$ predicts PDMS-related sub-scores.
The generator induces a trajectory distribution
$\pi_t(\tau\mid o)$.
The true expected score and scorer-surrogate expected score are
\begin{equation}
    J^*(\pi)
    =
    \mathbb{E}_{o\sim\mathcal{D},\,\tau\sim\pi(\cdot\mid o)}
    [R^*(o,\tau)],
\end{equation}
\begin{equation}
    J_{s_t}(\pi)
    =
    \mathbb{E}_{o\sim\mathcal{D},\,\tau\sim\pi(\cdot\mid o)}
    [s_t(o,\tau)].
\end{equation}

For a $K$-proposal set, we also use the empirical distribution
\begin{equation}
    \mu_t^o
    =
    \frac{1}{K}
    \sum_{i=1}^{K}
    \delta_{\tau_i^t(o)}.
\end{equation}

\subsection{Auxiliary Result: Approximate Monotonicity}
\label{app:proof_mono}

We first restate the assumptions.
Let the trust region around the current generator be
\begin{equation}
    \mathcal{B}_t(\rho)
    =
    \{\pi:d(\pi,\pi_t)\le\rho\}.
\end{equation}
Assume local scorer accuracy:
\begin{equation}
    \left|
    J_{s_t}(\pi)-J^*(\pi)
    \right|
    \le
    \epsilon_t(\rho),
    \qquad
    \forall \pi\in\mathcal{B}_t(\rho).
\end{equation}
Define
\begin{equation}
    e_t(\pi)
    =
    \left|
    J_{s_t}(\pi)-J^*(\pi)
    \right|,
\end{equation}
and the surrogate improvement
\begin{equation}
    m_t
    =
    J_{s_t}(\pi_{t+1})-J_{s_t}(\pi_t).
\end{equation}

By the definition of $e_t(\pi)$,
\begin{equation}
    J^*(\pi_{t+1})
    \ge
    J_{s_t}(\pi_{t+1})
    -
    e_t(\pi_{t+1}),
\end{equation}
and
\begin{equation}
    J^*(\pi_t)
    \le
    J_{s_t}(\pi_t)
    +
    e_t(\pi_t).
\end{equation}
Subtracting the second inequality from the first gives
\begin{align}
    J^*(\pi_{t+1})-J^*(\pi_t)
    &\ge
    J_{s_t}(\pi_{t+1})-J_{s_t}(\pi_t)
    -
    e_t(\pi_{t+1})
    -
    e_t(\pi_t) \\
    &=
    m_t-e_t(\pi_{t+1})-e_t(\pi_t).
\end{align}
This proves the general form.
Therefore, if
\begin{equation}
    m_t
    \ge
    e_t(\pi_{t+1})+e_t(\pi_t),
\end{equation}
then
\begin{equation}
    J^*(\pi_{t+1})\ge J^*(\pi_t).
\end{equation}

Now assume the generator update is conservative and remains in $\mathcal{B}_t(\rho)$.
If the generator phase improves the regularized surrogate up to residual $\alpha_t\ge0$:
\begin{equation}
    J_{s_t}(\pi_{t+1})
    -
    \lambda D(\pi_{t+1},\pi_t)
    \ge
    J_{s_t}(\pi_t)
    -
    \alpha_t,
\end{equation}
with $D(\pi_{t+1},\pi_t)\ge0$, then
\begin{equation}
    J_{s_t}(\pi_{t+1})
    \ge
    J_{s_t}(\pi_t)
    -
    \alpha_t,
\end{equation}
so $m_t\ge-\alpha_t$.
Since $\pi_t,\pi_{t+1}\in\mathcal{B}_t(\rho)$,
\begin{equation}
    e_t(\pi_t)\le\epsilon_t(\rho),
    \qquad
    e_t(\pi_{t+1})\le\epsilon_t(\rho).
\end{equation}
Substituting into the general bound yields
\begin{equation}
    J^*(\pi_{t+1})
    \ge
    J^*(\pi_t)
    -
    2\epsilon_t(\rho)
    -
    \alpha_t.
\end{equation}
This completes the proof.

\subsection{Auxiliary Result: One-step Scorer Extrapolation}
\label{app:proof_shift}

Define the scorer error function
\begin{equation}
    g_t(o,\tau)
    =
    |s_t(o,\tau)-R^*(o,\tau)|.
\end{equation}
Since both $s_t$ and $R^*$ lie in $[0,1]$, we have $0\le g_t\le1$.
Assume that the scorer phase fits the current generator distribution:
\begin{equation}
    \mathbb{E}_{\tau\sim\pi_t}
    [g_t(o,\tau)]
    \le
    \epsilon_t.
\end{equation}
Let the one-step drift be
\begin{equation}
    \Delta_t
    =
    \mathrm{TV}(\pi_{t+1},\pi_t).
\end{equation}
For any function $f\in[0,1]$,
\begin{equation}
    \left|
    \mathbb{E}_{P}[f]
    -
    \mathbb{E}_{Q}[f]
    \right|
    \le
    \mathrm{TV}(P,Q).
\end{equation}
Thus,
\begin{align}
    \mathbb{E}_{\tau\sim\pi_{t+1}}[g_t(o,\tau)]
    &\le
    \mathbb{E}_{\tau\sim\pi_t}[g_t(o,\tau)]
    +
    \mathrm{TV}(\pi_{t+1},\pi_t) \\
    &\le
    \epsilon_t+\Delta_t.
\end{align}
Summing over $t=0,\ldots,T-1$ gives
\begin{equation}
    \sum_{t=0}^{T-1}
    \mathbb{E}_{\pi_{t+1}}
    [|s_t-R^*|]
    \le
    \sum_{t=0}^{T-1}\epsilon_t
    +
    \sum_{t=0}^{T-1}\Delta_t.
\end{equation}
If $\epsilon_t\le\epsilon$ and $\Delta_t\le\rho$, the bound becomes
\begin{equation}
    \sum_{t=0}^{T-1}
    \mathbb{E}_{\pi_{t+1}}
    [|s_t-R^*|]
    \le
    T\epsilon+T\rho.
\end{equation}

We now compare this with using a fixed initial scorer $s_0$.
If
\begin{equation}
    \mathbb{E}_{\pi_0}
    [|s_0-R^*|]
    \le
    \epsilon_0,
\end{equation}
then
\begin{equation}
    \mathbb{E}_{\pi_{t+1}}
    [|s_0-R^*|]
    \le
    \epsilon_0+
    \mathrm{TV}(\pi_{t+1},\pi_0).
\end{equation}
By the triangle inequality,
\begin{equation}
    \mathrm{TV}(\pi_{t+1},\pi_0)
    \le
    \sum_{k=0}^{t}
    \Delta_k.
\end{equation}
Therefore,
\begin{equation}
    \mathbb{E}_{\pi_{t+1}}
    [|s_0-R^*|]
    \le
    \epsilon_0+
    \sum_{k=0}^{t}
    \Delta_k.
\end{equation}
Summing over $t$ yields
\begin{align}
    \sum_{t=0}^{T-1}
    \mathbb{E}_{\pi_{t+1}}
    [|s_0-R^*|]
    &\le
    T\epsilon_0+
    \sum_{t=0}^{T-1}
    \sum_{k=0}^{t}\Delta_k \\
    &=
    T\epsilon_0+
    \sum_{k=0}^{T-1}(T-k)\Delta_k.
\end{align}
If $\Delta_k\le\rho$, this is
\begin{equation}
    T\epsilon_0+\frac{T(T+1)}{2}\rho.
\end{equation}
Thus, refitting the scorer reduces the drift dependence from quadratic to linear in the number of refinement steps.

\subsection{Auxiliary Result: Approximate Pareto Consistency}
\label{app:proof_pareto}

Let the true vector score be
\begin{equation}
    \mathbf{R}^*(o,\tau)
    =
    (R_1^*(o,\tau),\ldots,R_M^*(o,\tau)),
\end{equation}
and the predicted vector score be
\begin{equation}
    \mathbf{s}_t(o,\tau)
    =
    (s_{t,1}(o,\tau),\ldots,s_{t,M}(o,\tau)).
\end{equation}
Assume uniform vector error on the candidate pool:
\begin{equation}
    \|\mathbf{s}_t(o,\tau)-\mathbf{R}^*(o,\tau)\|_{\infty}
    \le
    \delta_t,
    \qquad
    \forall \tau\in\mathcal{C}_t(o).
\end{equation}

We define the true $\epsilon$-Pareto set as
\begin{equation}
    \mathcal{P}^*_{\epsilon}(o)
    =
    \left\{
    \tau\in\mathcal{C}_t(o):
    \nexists \tau'\in\mathcal{C}_t(o)
    \text{ such that }
    R_m^*(o,\tau')
    \ge
    R_m^*(o,\tau)+\epsilon,\ \forall m
    \right\}.
\end{equation}
The scorer-induced Pareto set is
\begin{equation}
    \mathcal{P}_{s_t}(o)
    =
    \left\{
    \tau\in\mathcal{C}_t(o):
    \nexists \tau'\in\mathcal{C}_t(o)
    \text{ such that }
    s_{t,m}(o,\tau')
    \ge
    s_{t,m}(o,\tau),\ \forall m,
    \text{ and }
    \exists m:\ s_{t,m}(o,\tau')>s_{t,m}(o,\tau)
    \right\}.
\end{equation}

We prove that for any $\kappa>0$,
\begin{equation}
    \mathcal{P}_{s_t}(o)
    \subseteq
    \mathcal{P}^*_{2\delta_t+\kappa}(o).
\end{equation}
Suppose, for contradiction, that there exists
$\tau\in\mathcal{P}_{s_t}(o)$ but
$\tau\notin\mathcal{P}^*_{2\delta_t+\kappa}(o)$.
Then there exists $\tau'\in\mathcal{C}_t(o)$ such that
\begin{equation}
    R_m^*(o,\tau')
    \ge
    R_m^*(o,\tau)+2\delta_t+\kappa,
    \qquad
    \forall m.
\end{equation}
By the error bound,
\begin{equation}
    s_{t,m}(o,\tau')
    \ge
    R_m^*(o,\tau')-\delta_t,
\end{equation}
and
\begin{equation}
    s_{t,m}(o,\tau)
    \le
    R_m^*(o,\tau)+\delta_t.
\end{equation}
Combining these inequalities gives
\begin{align}
    s_{t,m}(o,\tau')
    &\ge
    R_m^*(o,\tau)+2\delta_t+\kappa-\delta_t \\
    &=
    R_m^*(o,\tau)+\delta_t+\kappa \\
    &\ge
    s_{t,m}(o,\tau)+\kappa.
\end{align}
Thus,
\begin{equation}
    s_{t,m}(o,\tau')>s_{t,m}(o,\tau),
    \qquad
    \forall m.
\end{equation}
Therefore $\tau'$ strictly dominates $\tau$ in scorer space, contradicting
$\tau\in\mathcal{P}_{s_t}(o)$.
This proves the result.

The additional $\kappa>0$ only excludes the degenerate boundary case where the true margin is exactly $2\delta_t$ and the scorer errors simultaneously attain their worst-case values in every objective.
Since $\kappa$ can be arbitrarily small, the result can be interpreted as containment in the right-limit $(2\delta_t)^+$-approximate Pareto region.

\paragraph{Coverage-error extension.}
Assume that the student generator covers each predicted Pareto target within distance $\eta_t$: for any $\tau\in\mathcal{P}_{s_t}(o)$, there exists a student proposal $\tilde{\tau}$ such that
\begin{equation}
    d_{\mathcal{X}}(\tilde{\tau},\tau)\le\eta_t.
\end{equation}
Assume also that $\mathbf{R}^*$ is $L$-Lipschitz on the valid proposal manifold:
\begin{equation}
    \|\mathbf{R}^*(o,\tau)-\mathbf{R}^*(o,\tau')\|_{\infty}
    \le
    Ld_{\mathcal{X}}(\tau,\tau').
\end{equation}
Then the covered student proposal lies in the true
$(2\delta_t+\kappa+L\eta_t)$-approximate Pareto region.
Indeed, if $\tilde{\tau}$ were dominated by some $\tau'$ with margin
$2\delta_t+\kappa+L\eta_t$, then
\begin{equation}
    R_m^*(o,\tau')
    \ge
    R_m^*(o,\tilde{\tau})
    +
    2\delta_t+\kappa+L\eta_t.
\end{equation}
Since
\begin{equation}
    R_m^*(o,\tilde{\tau})
    \ge
    R_m^*(o,\tau)-L\eta_t,
\end{equation}
we obtain
\begin{equation}
    R_m^*(o,\tau')
    \ge
    R_m^*(o,\tau)+2\delta_t+\kappa,
    \qquad
    \forall m,
\end{equation}
which contradicts
$\tau\in\mathcal{P}^*_{2\delta_t+\kappa}(o)$.

\subsection{Proof of Theorem~\ref{thm:enrichment}}
\label{app:proof_enrichment}

For a scene $o$, define the true high-score region
\begin{equation}
    \mathcal{H}_o
    =
    \{\tau:R^*(o,\tau)\ge r_{\mathrm{high}}\}.
\end{equation}
Let the current proposal distribution be
\begin{equation}
    \mu_t^o
    =
    \frac{1}{K}
    \sum_{i=1}^{K}
    \delta_{\tau_i^t(o)}.
\end{equation}
Let $A_t(o)$ denote the scorer-selected top-$k$ or vector-Pareto target set, and let
\begin{equation}
    \nu_t^o
    =
    \frac{1}{|A_t(o)|}
    \sum_{\tau\in A_t(o)}
    \delta_{\tau}
\end{equation}
be the empirical distribution on this target set.
Define
\begin{equation}
    p_t(o)=\mu_t^o(\mathcal{H}_o),
    \qquad
    q_t(o)=\nu_t^o(\mathcal{H}_o).
\end{equation}

The selected-set enrichment assumption states that
\begin{equation}
    q_t(o)\ge p_t(o)+\xi_t(o),
    \qquad
    \xi_t(o)>0.
\end{equation}
The conservative refinement assumption states that the actual student distribution is close to the ideal mixture
\begin{equation}
    \bar{\mu}_{t+1}^o
    =
    (1-\alpha_t)\mu_t^o+\alpha_t\nu_t^o,
    \qquad
    \alpha_t\in[0,1],
\end{equation}
in total variation:
\begin{equation}
    \mathrm{TV}
    \left(
    \mu_{t+1}^o,
    \bar{\mu}_{t+1}^o
    \right)
    \le
    \eta_t(o).
\end{equation}
Then
\begin{align}
    \bar{\mu}_{t+1}^o(\mathcal{H}_o)
    &=
    (1-\alpha_t)\mu_t^o(\mathcal{H}_o)
    +
    \alpha_t\nu_t^o(\mathcal{H}_o) \\
    &=
    (1-\alpha_t)p_t(o)+\alpha_t q_t(o) \\
    &\ge
    (1-\alpha_t)p_t(o)
    +
    \alpha_t[p_t(o)+\xi_t(o)] \\
    &=
    p_t(o)+\alpha_t\xi_t(o).
\end{align}
For any event $A$, total variation implies
\begin{equation}
    \mu_{t+1}^o(A)
    \ge
    \bar{\mu}_{t+1}^o(A)-\eta_t(o).
\end{equation}
Taking $A=\mathcal{H}_o$ yields
\begin{equation}
    p_{t+1}(o)
    =
    \mu_{t+1}^o(\mathcal{H}_o)
    \ge
    p_t(o)+\alpha_t\xi_t(o)-\eta_t(o).
\end{equation}
This proves the high-score support statement.

\paragraph{Expected-score version.}
Assume instead that the selected target distribution has higher true expected score:
\begin{equation}
    \mathbb{E}_{\nu_t^o}[R^*]
    -
    \mathbb{E}_{\mu_t^o}[R^*]
    \ge
    \beta_t(o)>0.
\end{equation}
Then
\begin{align}
    \mathbb{E}_{\bar{\mu}_{t+1}^o}[R^*]
    &=
    (1-\alpha_t)\mathbb{E}_{\mu_t^o}[R^*]
    +
    \alpha_t\mathbb{E}_{\nu_t^o}[R^*] \\
    &=
    \mathbb{E}_{\mu_t^o}[R^*]
    +
    \alpha_t
    \left(
    \mathbb{E}_{\nu_t^o}[R^*]
    -
    \mathbb{E}_{\mu_t^o}[R^*]
    \right) \\
    &\ge
    \mathbb{E}_{\mu_t^o}[R^*]
    +
    \alpha_t\beta_t(o).
\end{align}
Because $R^*\in[0,1]$ and
$\mathrm{TV}(\mu_{t+1}^o,\bar{\mu}_{t+1}^o)\le\eta_t(o)$,
\begin{equation}
    \mathbb{E}_{\mu_{t+1}^o}[R^*]
    \ge
    \mathbb{E}_{\bar{\mu}_{t+1}^o}[R^*]
    -
    \eta_t(o).
\end{equation}
Thus,
\begin{equation}
    \mathbb{E}_{\mu_{t+1}^o}[R^*]
    \ge
    \mathbb{E}_{\mu_t^o}[R^*]
    +
    \alpha_t\beta_t(o)
    -
    \eta_t(o).
\end{equation}

\paragraph{Multi-round extension.}
Summing over $t=0,\ldots,T-1$ gives
\begin{equation}
    \mathbb{E}_{\mu_T^o}[R^*]
    \ge
    \mathbb{E}_{\mu_0^o}[R^*]
    +
    \sum_{t=0}^{T-1}\alpha_t\beta_t(o)
    -
    \sum_{t=0}^{T-1}\eta_t(o).
\end{equation}
Therefore, Stage-2 refinement improves the true expected proposal quality whenever
\begin{equation}
    \sum_{t=0}^{T-1}\alpha_t\beta_t(o)
    >
    \sum_{t=0}^{T-1}\eta_t(o).
\end{equation}

\paragraph{From high-score support to Oracle@$K$.}
If $K$ proposals are interpreted as samples from $\mu_t^o$, the probability that at least one proposal falls in the high-score region is
\begin{equation}
    A_K(p_t)
    =
    1-(1-p_t)^K.
\end{equation}
If all trajectories in $\mathcal{H}_o$ have score at least $r_{\mathrm{high}}$ and all candidates have score at least $r_{\min}$, then the oracle score satisfies
\begin{equation}
    O_t(o)
    \ge
    r_{\min}
    +
    (r_{\mathrm{high}}-r_{\min})A_K(p_t).
\end{equation}
Since $A_K(p)$ is monotone in $p$, increasing $p_t(o)$ increases this lower bound on Oracle@$K$.

\paragraph{From Oracle@$K$ to deployed top-1.}
Let
\begin{equation}
    O_t(o)
    =
    \max_{\tau\in T_t(o)} R^*(o,\tau)
\end{equation}
be the true oracle in the proposal set, and let
\begin{equation}
    V_t(o)
    =
    R^*
    \left(
    o,
    \arg\max_{\tau\in T_t(o)}
    s_t(o,\tau)
    \right)
\end{equation}
be the true score of the deployed scorer-selected top-1.
Define ranking regret
\begin{equation}
    G_t(o)=O_t(o)-V_t(o).
\end{equation}
Then
\begin{equation}
    V_T(o)-V_0(o)
    =
    [O_T(o)-O_0(o)]
    +
    [G_0(o)-G_T(o)].
\end{equation}
Thus, deployed top-1 improvement comes from two sources:
improving the proposal oracle and reducing ranking regret.

\subsection{Uniform Margin as a Sufficient Condition}
\label{app:uniform_margin}

The selected-set enrichment assumption in Theorem~\ref{thm:enrichment} is weaker than worst-case scorer calibration.
A clean sufficient condition is the standard margin condition.

Assume there are high-score and low-score regions $\mathcal{H}_o$ and $\mathcal{L}_o$ such that
\begin{equation}
    R^*(o,\tau_h)\ge r_{\mathrm{high}},
    \quad
    \forall \tau_h\in\mathcal{H}_o,
\end{equation}
\begin{equation}
    R^*(o,\tau_l)\le r_{\mathrm{low}},
    \quad
    \forall \tau_l\in\mathcal{L}_o,
\end{equation}
with margin
\begin{equation}
    \gamma=r_{\mathrm{high}}-r_{\mathrm{low}}>0.
\end{equation}
If the scorer satisfies the uniform error bound
\begin{equation}
    |s_t(o,\tau)-R^*(o,\tau)|\le\epsilon,
    \qquad
    \forall \tau\in\mathcal{H}_o\cup\mathcal{L}_o,
\end{equation}
and $\gamma>2\epsilon$, then for any
$\tau_h\in\mathcal{H}_o$ and $\tau_l\in\mathcal{L}_o$,
\begin{align}
    s_t(o,\tau_h)
    &\ge
    R^*(o,\tau_h)-\epsilon
    \ge
    r_{\mathrm{high}}-\epsilon, \\
    s_t(o,\tau_l)
    &\le
    R^*(o,\tau_l)+\epsilon
    \le
    r_{\mathrm{low}}+\epsilon.
\end{align}
Therefore,
\begin{equation}
    s_t(o,\tau_h)-s_t(o,\tau_l)
    \ge
    \gamma-2\epsilon>0.
\end{equation}
Thus high-score proposals are ranked above low-score proposals.

This condition is sufficient but not necessary.
In practice, the worst-case error $\epsilon_{\max}$ can be overly conservative.
Stage-2 distillation only requires the weaker statistical condition that scorer-selected top-$k$ or Pareto targets are enriched with high true evaluator scores.

\subsection{Empirical Support for Selected-Set Enrichment}
\label{app:enrichment_empirical}

We empirically validate the selected-set enrichment assumption used in Theorem~\ref{thm:enrichment}.
The analysis is conducted on the NAVSIM evaluation PKL with 12,146 scenes and 64 proposals per scene.

Table~\ref{tab:selected_set_enrichment} summarizes the empirical proxies.
The worst-case margin condition $\gamma>2\epsilon_{\max}$ is intentionally conservative and is not expected to hold for all proposals.
Indeed, under the per-scene elite/reject $(0.05/0.05)$ protocol, only $10.64\%$ of scenes satisfy $\gamma_o>2\epsilon_{\max,o}$.
However, the weaker statistical conditions needed by Stage-2 distillation are strongly supported.
Under elite/reject $(0.01/0.01)$, $92.50\%$, $82.37\%$, and $71.56\%$ of scenes satisfy
$\gamma_o>2\epsilon_{p75,o}$,
$\gamma_o>2\epsilon_{p90,o}$,
and
$\gamma_o>2\epsilon_{p95,o}$, respectively.
Moreover, fixed-threshold high/low groups are highly separable: for high $\ge0.95$ and low $\le0.50$, pairwise ranking accuracy reaches $94.72\%$.
Most importantly, scorer-high proposals are truly high scoring:
when $s\ge0.95$, the true mean score is $0.9753$,
$P(R^*\ge0.90)=96.69\%$, and
$P(R^*=1)=69.74\%$, compared with a pooled full-score rate of $35.42\%$.
These observations support the key premise of Theorem~\ref{thm:enrichment}: Stage 2 improves over Stage 1 not by discovering high-quality modes from scratch, but by reallocating proposal mass toward scorer-selected targets that are statistically enriched with high true evaluator scores.

\begin{table}[t]
\centering
\small
\caption{
\textbf{Empirical validation of selected-set enrichment.}
Worst-case scorer calibration is too conservative, while high-probability separability, high-score precision, and scorer Top-$k$ coverage support the enrichment condition required by Stage-2 distillation.
}
\label{tab:selected_set_enrichment}
\resizebox{\linewidth}{!}{
\begin{tabular}{llcl}
\toprule
Validation target & Metric / protocol & Result & Interpretation \\
\midrule
Stage-1 support
& Pooled full-score proposals
& $35.42\%$
& Stage 1 provides non-zero high-score support. \\

Worst-case margin
& $P(\gamma_o>2\epsilon_{\max,o})$, elite/reject $0.05/0.05$
& $10.64\%$
& Worst-case calibration is too strict. \\

High-prob. margin
& $P(\gamma_o>2\epsilon_{p75,o})$, elite/reject $0.01/0.01$
& $92.50\%$
& Quantile-level separation is widely satisfied. \\

High-prob. margin
& $P(\gamma_o>2\epsilon_{p90,o})$, elite/reject $0.01/0.01$
& $82.37\%$
& Most scenes satisfy high-probability separation. \\

High-prob. margin
& $P(\gamma_o>2\epsilon_{p95,o})$, elite/reject $0.01/0.01$
& $71.56\%$
& Strict quantile-level separation still often holds. \\

Pairwise separability
& high $\ge0.95$, low $\le0.50$
& $94.72\%$
& Scorer usually ranks true high-score proposals above low-score ones. \\

High-score precision
& $s\ge0.90$
& mean $R^*=0.9529$, $P(R^*\ge0.90)=90.16\%$
& Scorer-high proposals are reliable true high-score candidates. \\

High-score precision
& $s\ge0.95$
& mean $R^*=0.9753$, $P(R^*\ge0.90)=96.69\%$, $P(R^*=1)=69.74\%$
& High-confidence scorer targets are strongly enriched. \\

Top-$k$ quality
& scorer Top-1
& mean best $R^*=0.9448$, oracle gap $0.0495$
& Single top selection is already high quality. \\

Top-$k$ quality
& scorer Top-8
& mean best $R^*=0.9656$, oracle gap $0.0283$
& Top-$k$ targets improve high-score coverage. \\

Top-$k$ quality
& scorer Top-16
& mean best $R^*=0.9730$, oracle gap $0.0209$
& Larger target sets further reduce oracle gap. \\
\bottomrule
\end{tabular}
}
\end{table}

At the pooled level, let $\mathcal{H}=\{\tau:R^*(\tau)=1\}$.
The proposal pool has
\begin{equation}
    p \approx 35.42\%.
\end{equation}
Using $s\ge0.95$ as a high-confidence scorer-selected proxy gives
\begin{equation}
    q \approx 69.74\%.
\end{equation}
Thus the enrichment gap is approximately
\begin{equation}
    q-p
    \approx
    69.74\%-35.42\%
    =
    34.32\%.
\end{equation}
This directly supports the condition
$q_t(o)\ge p_t(o)+\xi_t(o)$
in an aggregate sense.

\section{Implementation Details and Hyperparameters}
\label{app:implementation_details}

This appendix summarizes the main implementation details and hyperparameters used in CLOVER. 
Unless otherwise specified, all models and ablations use the same DrivoR-style base architecture shown in Table~\ref{tab:base_architecture_hparams}. 
Stage-1 diversity pretraining and its loss weights are summarized in Table~\ref{tab:stage1_hparams}. 
The pseudo-expert generation configuration used for evaluator-filtered trajectory coverage is provided in Table~\ref{tab:pseudo_expert_generation_hparams}. 
Stage-2 alternating self-distillation and its teacher-set configuration are summarized in Table~\ref{tab:stage2_hparams}. 
Finally, Table~\ref{tab:compute_resources} reports the compute resources and wall-clock training time. 

\begin{table}[t]
\centering
\small
\caption{
\textbf{Base DrivoR-style architecture used by CLOVER.}
Unless otherwise specified, all models and ablations use this common architecture.
}
\label{tab:base_architecture_hparams}
\begin{tabular}{ll}
\toprule
Hyperparameter & Value \\
\midrule
Input cameras & front, left, right, back \\
LiDAR input & disabled \\
Backbone & DINOv2 ViT-S \\
Backbone identifier & \texttt{timm/vit\_small\_patch14\_reg4\_dinov2.lvd142m} \\
LoRA & enabled, rank 32 \\
Scene tokens per camera & 16 \\
Prediction horizon & 4.0 s \\
Sampling interval & 0.5 s \\
Number of future poses & 8 \\
Number of proposals & 64 \\
Transformer hidden dimension & 256 \\
Transformer FFN dimension & 1024 \\
Trajectory representation & $(x,y,\theta)$ \\
One token per trajectory & true \\
NAVSIM v1 deployment weights & NC/DAC/DDC/TTC/EP/Comfort = 1/1/0/5/5/2 \\
\bottomrule
\end{tabular}
\end{table}

\begin{table}[t]
\centering
\small
\caption{
\textbf{Stage-1 diversity pretraining hyperparameters.}
Stage 1 trains the generator with logged-trajectory imitation and evaluator-filtered pseudo-expert coverage, while pretraining the scorer with evaluator-provided trajectory scores.
}
\label{tab:stage1_hparams}
\begin{tabular}{ll}
\toprule
Hyperparameter & Value \\
\midrule
Training epochs & 25 \\
Batch size & 32 \\
Number of GPUs & 4 \\
DataLoader workers & 4 \\
Optimizer & AdamW \\
Base learning rate & $2\times10^{-4}$ \\
Precision & 16-mixed \\
Seed & 2 \\
Ray scoring & enabled \\
Ray threads per node & 96 \\
Pseudo-expert top-$K$ & 8 \\
Pseudo-expert score threshold & 0.8 \\
Pseudo-expert loss weight & 0.5 \\
Pseudo-expert applied to all refinement layers & false \\
\midrule
Trajectory imitation loss weight & 1.0 \\
Final score loss weight & 1.0 \\
Prediction cross-entropy loss weight & 1.0 \\
Prediction L1 loss weight & 0.1 \\
Prediction area loss weight & 2.0 \\
\bottomrule
\end{tabular}
\end{table}

\begin{table}[t]
\centering
\small
\caption{
\textbf{Pseudo-expert trajectory generation hyperparameters.}
Pseudo-expert trajectories are generated from interpretable candidate families and filtered by evaluator-based trajectory quality.
}
\label{tab:pseudo_expert_generation_hparams}
\begin{tabular}{ll}
\toprule
Hyperparameter & Value \\
\midrule
Number of poses & 8 \\
Sampling interval & 0.5 s \\
Speed candidates & $(0,2,4,6,8,10,12,15)$ m/s \\
Regular lateral offsets & $(-3.5,-2.0,-1.0,-0.5,0,0.5,1.0,2.0,3.5)$ m \\
Off-road lateral offsets & $(-7.0,-5.5,5.5,7.0)$ m \\
Transition portions & $(0.35,0.6,1.0)$ \\
Acceleration rates & $(-2.0,-1.0,-0.5,0.5,1.0,2.0)$ \\
Score bins & $(0.0,0.2,0.4,0.6,0.8,1.01)$ \\
Progress bins & $(0.0,0.2,0.5,0.8,1.01)$ \\
Retained trajectories per scene & 50 \\
Maximum scored candidates per scene & 180 \\
\bottomrule
\end{tabular}
\end{table}

\begin{table}[t]
\centering
\small
\caption{
\textbf{Stage-2 alternating self-distillation hyperparameters.}
Stage 2 alternates between scorer fitting and generator refinement. 
The visual backbone is frozen during Stage 2, while the scorer remains trainable.
}
\label{tab:stage2_hparams}
\begin{tabular}{ll}
\toprule
Hyperparameter & Value \\
\midrule
Initialization & Stage-1 checkpoint \\
Number of cycles & 30 \\
Scorer-fitting epochs per cycle & 1 \\
Generator-refinement epochs per cycle & 1 \\
Total epochs & 60 \\
Batch size & 32 \\
Number of GPUs & 4 \\
DataLoader workers & 8 \\
Base learning rate & $3\times10^{-5}$ \\
Starting phase & scorer fitting \\
Freeze visual backbone & true \\
Detach proposals in scorer & true \\
Sanitize proposals & true \\
Proposal XY limit & 100.0 \\
Ray scoring & enabled \\
Ray threads per node & 96 \\
\midrule
Trajectory loss weight & 0.1 \\
Intermediate loss weight & 0.02 \\
Final score loss weight & 1.0 \\
Pareto guidance weight & 1.0 \\
Teacher stability weight & 0.05 \\
Pareto set maximum size & 8 \\
Pareto set minimum size & 2 \\
Teacher reward threshold & 0.4 \\
\bottomrule
\end{tabular}
\end{table}

\begin{table}[t]
\centering
\small
\caption{
\textbf{Compute resources and training time.}
Training time includes real-time trajectory scoring with the NAVSIM evaluator, which depends on CPU throughput in addition to GPU training speed.
}
\label{tab:compute_resources}
\begin{tabular}{ll}
\toprule
Item & Value \\
\midrule
GPUs & 4 NVIDIA A100 \\
CPU & AMD EPYC 7742 64-Core Processor \\
Stage-1 training time & 30--40 hours \\
Stage-2 training time & 60--80 hours \\
Total training time & 4--5 days \\
\bottomrule
\end{tabular}
\end{table}

\section{Fixed-Proposal Scorer Diagnostic}
\label{app:scorer_diagnostic}

The proposal diversity analysis shows that CLOVER substantially improves both proposal quality and proposal diversity. 
However, it also reveals a new bottleneck: once many high-quality candidates are available, final performance depends heavily on whether the scorer can correctly rank near-tied but behaviorally different trajectories. 
A better scorer can improve the final selected trajectory directly, and in our framework it can also serve as a differentiable mediator for refining the generator during self-distillation. 
This motivates a standardized diagnostic protocol that focuses specifically on trajectory-level scorer design.

We introduce a fixed-proposal scorer diagnostic protocol under a shared CLOVER-generated candidate pool. 
For each scene in NAVSIM \texttt{navtrain} and \texttt{navtest}, we use the trained CLOVER generator to produce 64 candidate trajectories. 
Each candidate is then evaluated by the true PDMS evaluator to obtain scalar scores and interpretable sub-score labels. 
This produces a scorer-only dataset:
\begin{equation}
    \mathcal{D}_{\mathrm{score}}
    =
    \left\{
    \left(
    o,
    \{\tau_i\}_{i=1}^{64},
    \{\mathbf{r}^{\mathrm{true}}_i\}_{i=1}^{64}
    \right)
    \right\}.
\end{equation}
A scorer architecture is trained to predict sub-scores for each fixed candidate and selects the trajectory with the highest composed score:
\begin{equation}
    i^\ast_{\phi}
    =
    \arg\max_i
    \Gamma_{\mathrm{PDMS}}
    \left(
    S_{\phi}(o,\tau_i)
    \right).
\end{equation}
We report the true PDMS of the selected trajectory:
\begin{equation}
    \mathrm{Score}
    =
    \frac{1}{|\mathcal{D}|}
    \sum_{o \in \mathcal{D}}
    s^{\mathrm{true}}_{i^\ast_{\phi}}.
\end{equation}

This protocol isolates scorer quality from proposal generation, because all scorer architectures are evaluated under the same candidate trajectories and the same true labels. 
It is lightweight, reproducible, and directly targets the ranking component of proposal-selection planners. 
We deliberately treat it as a scorer diagnostic rather than a full planner benchmark: it does not include generator-scorer closed-loop training, and the fixed offline candidate pool differs from a fully deployed CLOVER system. 
Therefore, its scores should not be compared directly with full-system PDMS results.

Table~\ref{tab:scorer_diagnostic_backbones} compares several scorer backbones under this protocol. 
DINOv2-Small is the backbone used in our main CLOVER model. 
Larger DINO backbones provide modest test improvements, while the Wan2.2-5B video-generation backbone achieves the highest diagnostic test score. 
For Wan2.2-5B, we LoRA fine-tune the video model on NAVSIM and use intermediate block-15 features for scoring. 
This result is consistent with the intuition that video or world-model features may better capture future-scene-dependent factors required by PDMS. 
However, Wan2.2-5B also shows a clear train-test gap and is far more expensive than the DINOv2-Small scorer. 
Similarly, DINO base variants increase training cost in our closed-loop refinement pipeline. 
For these reasons, we keep DINOv2-Small as the deployed scorer in CLOVER and use the larger backbones only as diagnostic references.

\begin{table}[t]
\centering
\small
\caption{
\textbf{Fixed-proposal scorer diagnostic.}
All scorers are trained and evaluated on the same CLOVER-generated 64-candidate pools with true PDMS sub-score labels. 
Train and test scores denote the true PDMS of the scorer-selected top-1 trajectory on \texttt{navtrain} and \texttt{navtest}, respectively. 
These diagnostic scores evaluate scorer ranking quality under fixed proposals and are not directly comparable to full-system CLOVER results.
}
\label{tab:scorer_diagnostic_backbones}
\begin{tabular}{lcc}
\toprule
Scorer backbone & Train score $\uparrow$ & Test score $\uparrow$ \\
\midrule
DINOv2-Small & 95.0 & 92.4 \\
DINOv2-Base & 95.3 & 92.6 \\
DINOv3-Small & 95.1 & 92.4 \\
DINOv3-Base & 95.1 & 92.7 \\
Wan2.2-5B features & \textbf{97.7} & \textbf{92.8} \\
\bottomrule
\end{tabular}
\end{table}

The diagnostic results suggest two observations. 
First, scorer design has measurable impact even when the candidate pool is fixed, confirming that trajectory ranking is a meaningful bottleneck after proposal diversity has been improved. 
Second, stronger visual or video features can improve scorer performance, but the gain must be balanced against computational cost and overfitting risk. 
The large train-test gap of Wan2.2-5B indicates high representational capacity, but also suggests that using such a model as an online scorer inside the full CLOVER training loop would be impractical under our current training budget.
We therefore release this scorer diagnostic protocol to support future studies on trajectory-level scoring, score calibration, and ranking architectures under a shared high-quality candidate pool.

\section{Seed Stability}
\label{app:seed_stability}

The official NAVSIM evaluation is deterministic given a trained model and a fixed benchmark split.
Therefore, the main source of stochastic variation in our reported results comes from training randomness, including parameter initialization, dataloader order, and stochastic optimization.
To assess this effect, we repeat the main CLOVER training with three random seeds under the same architecture, data split, hyperparameters, and evaluation protocol.

Table~\ref{tab:seed_stability} reports the seed-level variation.
Across three runs, the final NAVSIM v1 PDMS differs by less than $0.02$ points.
This variation is below the one-decimal reporting precision used in the main benchmark tables and does not affect any conclusion or ranking trend.
We therefore report the official single-run benchmark scores in the main paper and use this experiment only to verify run-to-run stability.

\begin{table}[h]
\centering
\small
\caption{
\textbf{Seed stability of CLOVER.}
All runs use the same architecture, training schedule, data split, and evaluation protocol.
The observed variation is below the one-decimal reporting precision of the main benchmark tables.
}
\label{tab:seed_stability}
\begin{tabular}{lcccc}
\toprule
Setting & Seed 1 & Seed 2 & Seed 3 & Max--Min \\
\midrule
NAVSIM v1 PDMS & 94.488 & 94.502 & 94.492 & $<0.02$ \\
NAVSIM v2 EPDMS & 90.354 & 90.358 & 90.368 & $<0.01$ \\
NAVSIM v2 EPDMS$^\ast$ & 87.201 & 87.207 & 87.210 & $<0.01$ \\
\bottomrule
\end{tabular}
\end{table}

\section{Additional Proposal Diversity Metrics}
\label{app:proposal_diversity}

We provide additional proposal-set statistics comparing the DrivoR baseline, Stage 1 diversity pretraining, and Stage 2 self-distillation refinement.
All metrics are averaged over 12,146 common scenes, with 64 proposals per scene.
The true PDMS scores are used only for analysis.

\paragraph{Quality metrics.}
Selected PDMS is the true PDMS score of the trajectory selected by the learned scorer.
Oracle@64 is the best true PDMS score among all 64 proposals.
Mean PDMS over 64 measures the average quality of the full proposal set.
Gap is the difference between Oracle@64 and Selected PDMS.
Count(PDMS$>0.95$), Count(PDMS$>0.90$), and Count(PDMS$<0.50$) summarize the proposal score distribution.

\paragraph{Diversity metrics.}
Pairwise ADE/FDE measure the average pairwise distance between generated proposals.
Endpoint Std Radius and Endpoint Area measure the spread of final trajectory endpoints.
Trajectory Effective Rank is computed by flattening each 8-step $(x,y)$ trajectory into a vector, centering the proposal set, and computing the entropy-based effective rank of the singular-value spectrum.
Endpoint Cluster Count@r greedily counts endpoint clusters under radius $r$.

\paragraph{Quality-aware diversity.}
To avoid overestimating diversity due to low-quality outliers, we also report diversity after filtering proposals with PDMS$\geq0.8$.
Qualified Cluster Count@2m measures the number of distinct endpoint clusters among qualified proposals.
Qualified Pairwise ADE/FDE measures diversity within qualified proposals.
Top-6 Real Pairwise ADE/FDE measures diversity among the six proposals with the highest true PDMS scores in each scene.

\begin{table*}[t]
\centering
\small
\caption{
\textbf{Top-K proposal quality under scorer ranking.}
TopK-Oracle@K reports the best true PDMS score among the scorer-ranked top-$K$ proposals. 
TopK-Mean@K reports the average true PDMS score within the scorer-ranked top-$K$ proposals.
}
\label{tab:topk_stage_analysis}
\begin{tabular}{lcccccc}
\toprule
Metric & $K$ & Baseline & Stage 1 & Stage 2 & Stage1 - Base & Stage2 - Stage1 \\
\midrule
TopK-Oracle@K & 1 & 0.9369 & 0.9413 & \textbf{0.9443} & +0.0043 & +0.0031 \\
TopK-Oracle@K & 2 & 0.9461 & 0.9514 & \textbf{0.9517} & +0.0054 & +0.0002 \\
TopK-Oracle@K & 3 & 0.9512 & \textbf{0.9570} & 0.9554 & +0.0058 & -0.0017 \\
TopK-Oracle@K & 6 & 0.9603 & \textbf{0.9660} & 0.9625 & +0.0057 & -0.0035 \\
\midrule
TopK-Mean@K & 1 & 0.9369 & 0.9413 & \textbf{0.9443} & +0.0043 & +0.0031 \\
TopK-Mean@K & 2 & 0.9353 & 0.9402 & \textbf{0.9435} & +0.0049 & +0.0033 \\
TopK-Mean@K & 3 & 0.9345 & 0.9383 & \textbf{0.9430} & +0.0039 & +0.0046 \\
TopK-Mean@K & 6 & 0.9320 & 0.9346 & \textbf{0.9416} & +0.0026 & +0.0070 \\
\bottomrule
\end{tabular}
\end{table*}

\begin{table*}[t]
\centering
\small
\caption{
\textbf{Full proposal-set quality statistics.}
Stage 1 achieves the highest oracle upper bound but also introduces a lower-quality tail. 
Stage 2 refines the expanded proposal distribution and achieves the best selected score, smallest oracle-selection gap, highest proposal mean, and fewest low-score proposals.
}
\label{tab:proposal_quality_full}
\begin{tabular}{lcccccc}

\toprule
Metric & Baseline & Stage 1 & Stage 2 & Stage1 - Base & Stage2 - Stage1 & Stage2 - Base \\
\midrule
Selected PDMS $\uparrow$ & 0.9369 & 0.9413 & \textbf{0.9443} & +0.0043 & +0.0031 & +0.0074 \\
Oracle@64 PDMS $\uparrow$ & 0.9933 & \textbf{0.9976} & 0.9939 & +0.0042 & -0.0037 & +0.0005 \\
Gap $\downarrow$ & 0.0564 & 0.0563 & \textbf{0.0495} & -0.0001 & -0.0068 & -0.0069 \\
Mean PDMS over 64 $\uparrow$ & 0.7972 & 0.7374 & \textbf{0.8277} & -0.0597 & +0.0903 & +0.0306 \\
Std PDMS over 64 & 0.2152 & \textbf{0.3013} & 0.2235 & +0.0861 & -0.0778 & +0.0084 \\
Count(PDMS$>0.95$) $\uparrow$ & 30.94 & 24.78 & \textbf{31.97} & -6.16 & +7.19 & +1.03 \\
Count(PDMS$>0.90$) $\uparrow$ & \textbf{40.24} & 32.13 & 39.90 & -8.10 & +7.77 & -0.33 \\
Count(PDMS$<0.50$) $\downarrow$ & 9.05 & 12.21 & \textbf{6.83} & +3.16 & -5.39 & -2.22 \\
\bottomrule
\end{tabular}
\end{table*}

\begin{table*}[t]
\centering
\small
\caption{
\textbf{Full proposal diversity statistics.}
Stage 1 produces the widest raw proposal spread, while Stage 2 slightly contracts this spread but remains substantially more diverse than the baseline. 
Higher is better for all metrics.
}
\label{tab:proposal_diversity_full}
\begin{tabular}{lcccccc}
\toprule
Metric & Baseline & Stage 1 & Stage 2 & Stage1 - Base & Stage2 - Stage1 & Stage2 - Base \\
\midrule
Pairwise ADE & 1.8004 & \textbf{5.9724} & 5.1956 & +4.1720 & -0.7768 & +3.3952 \\
Pairwise FDE & 4.4660 & \textbf{10.6973} & 8.2536 & +6.2313 & -2.4438 & +3.7876 \\
Endpoint Std Radius & 3.9033 & \textbf{10.2638} & 7.9477 & +6.3604 & -2.3160 & +4.0444 \\
Endpoint Area & 5.7837 & \textbf{46.1758} & 25.5615 & +40.3921 & -20.6143 & +19.7777 \\
Trajectory Effective Rank & 1.1401 & 1.2362 & \textbf{1.2748} & +0.0962 & +0.0385 & +0.1347 \\
\bottomrule
\end{tabular}
\end{table*}

\begin{table*}[t]
\centering
\small
\caption{
\textbf{Quality-aware proposal diversity.}
Diversity is computed either after filtering proposals with PDMS$\geq0.8$ or within the six highest-scoring proposals according to the true evaluator. 
This avoids attributing diversity gains only to low-quality outliers.
}
\label{tab:quality_aware_diversity}
\begin{tabular}{lcccccc}
\toprule
Metric & Baseline & Stage 1 & Stage 2 & Stage1 - Base & Stage2 - Stage1 & Stage2 - Base \\
\midrule
Qualified Count(PDMS$\geq0.8$) & 50.16 & 43.97 & \textbf{50.44} & -6.19 & +6.47 & +0.27 \\
Qualified Cluster Count@2m & 6.02 & \textbf{10.91} & 8.71 & +4.90 & -2.20 & +2.70 \\
Qualified Pairwise ADE & 1.5576 & \textbf{3.7988} & 3.6727 & +2.2412 & -0.1261 & +2.1151 \\
Qualified Pairwise FDE & 3.8024 & \textbf{6.8121} & 5.6674 & +3.0097 & -1.1447 & +1.8650 \\
Top-6 Real Pairwise ADE & 0.9110 & \textbf{2.1111} & 1.7507 & +1.2001 & -0.3603 & +0.8398 \\
Top-6 Real Pairwise FDE & 2.1955 & \textbf{3.5542} & 2.5910 & +1.3587 & -0.9633 & +0.3954 \\
\bottomrule
\end{tabular}
\end{table*}

\begin{table}[t]
\centering
\small
\caption{
\textbf{Endpoint cluster count under different radii.}
Endpoint clusters are computed by greedy clustering over final trajectory endpoints. 
Stage 1 produces the most endpoint branches, while Stage 2 remains consistently more diverse than the baseline at all radii.
}
\label{tab:endpoint_cluster_stage}
\begin{tabular}{lccc}
\toprule
Radius & Baseline & Stage 1 & Stage 2 \\
\midrule
1m & 16.77 & \textbf{32.77} & 23.20 \\
2m & 8.07 & \textbf{19.81} & 13.81 \\
3m & 5.29 & \textbf{14.65} & 9.88 \\
4m & 3.97 & \textbf{11.24} & 8.07 \\
\bottomrule
\end{tabular}
\end{table}

\paragraph{Interpretation.}
The three stages exhibit distinct behavior. 
Stage 1 is an expansion stage: it increases the oracle upper bound and raw geometric diversity, but also introduces more low-score proposals. 
Stage 2 is a refinement stage: it preserves much of the expanded support while improving scorer-selected top-1, top-$K$ mean quality, full-set mean quality, and low-score tail suppression. 
This supports the design of CLOVER as a two-stage process: first broaden the proposal space, then refine it toward high-value regions.

\subsection{Additional Qualitative Comparisons of Proposal Diversity}
\label{app:proposal_diversity_qualitative}

\begin{figure*}[p]
    \centering
    \includegraphics[width=\textwidth,keepaspectratio]{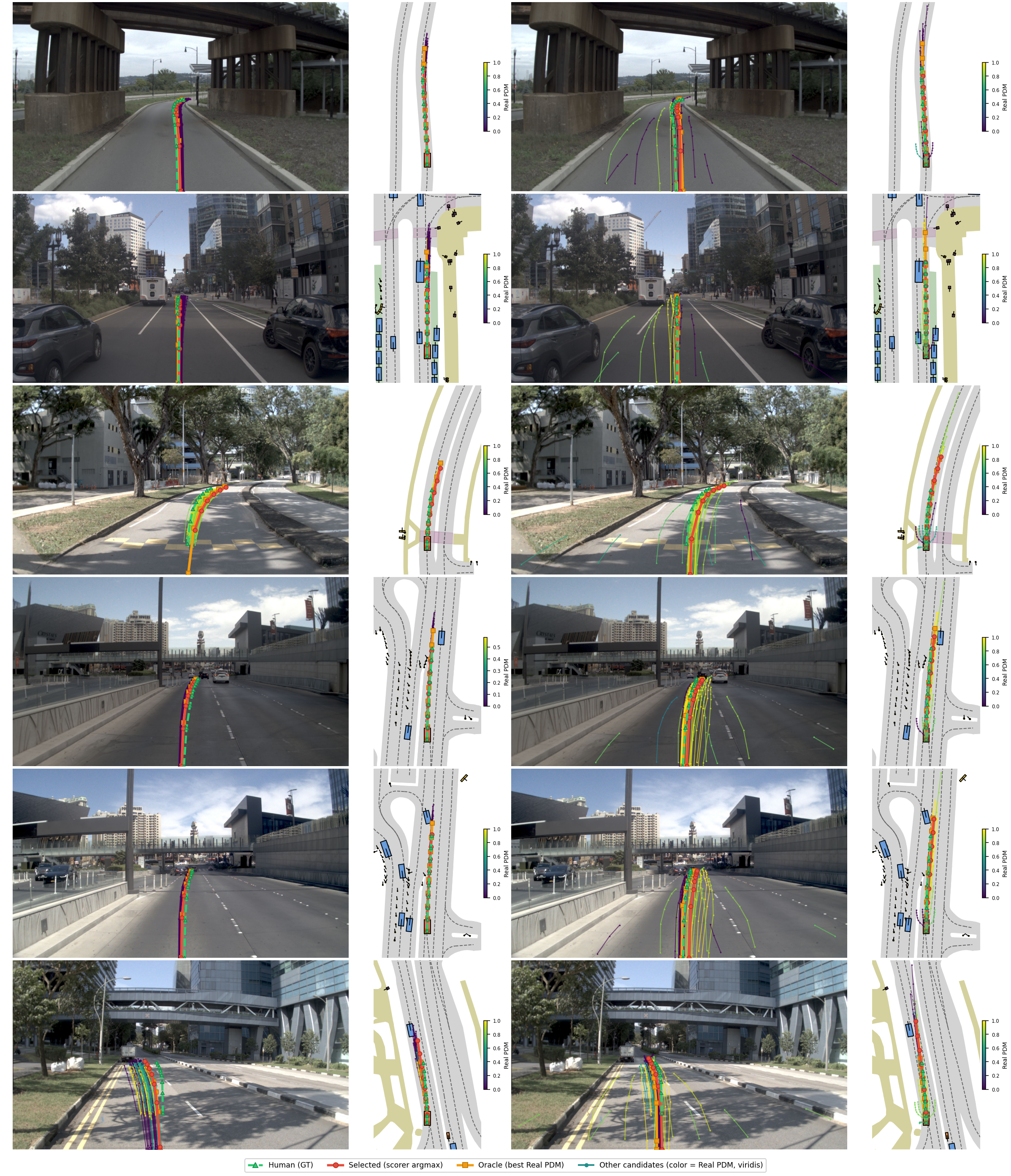}
    \caption{
    \textbf{Additional qualitative comparisons of proposal diversity on six scenes.}
    In each row, the left two panels show the DrivoR baseline and the right two panels show CLOVER after Stage-2 refinement. 
    For each method, we visualize all 64 candidate trajectories in the front-view image and in bird's-eye view. 
    Candidate trajectories are colored by their true PDMS scores for analysis. 
    Green triangles denote the human trajectory, red circles denote the scorer-selected trajectory, orange squares denote the oracle trajectory with the highest true PDMS, and the remaining candidates are colored by real PDMS using the viridis colormap. 
    Across all six scenes, the baseline proposals are concentrated around a narrow mode, whereas CLOVER consistently covers a wider set of feasible and often high-scoring candidates.
    }
    \label{fig:proposal_diversity_qualitative}
\end{figure*}

Figure~\ref{fig:proposal_diversity_qualitative} presents additional qualitative comparisons between the DrivoR baseline and CLOVER on six more scenes, following the same visualization protocol as Figure~\ref{fig:diversity_visualization} in the main text. 
For each scene, the left two panels show the DrivoR baseline and the right two panels show CLOVER after Stage-2 refinement. 
For each method, we visualize all 64 generated candidate trajectories in both the front-view image and bird's-eye view. 
Candidate trajectories are colored by their true PDMS scores for analysis, while the human trajectory, the scorer-selected trajectory, and the oracle trajectory with the highest true PDMS are highlighted explicitly.

These additional examples consistently support the same conclusion as in the main text. 
The DrivoR baseline tends to concentrate proposals around a narrow single mode, often with limited lateral or longitudinal variation. 
In contrast, CLOVER produces a broader set of feasible candidates, covering richer within-lane offsets, progress patterns, and alternative motion branches. 
Importantly, this increased diversity is not merely caused by low-quality outliers: many diverse candidates remain high-scoring under the evaluator, and the scorer-selected trajectory typically remains close to the oracle high-score mode. 
Together with the quantitative results in Section~\ref{sec:proposal_diversity}, these visualizations further confirm that CLOVER improves proposal diversity while preserving strong proposal quality.

%%%%%%%%%%%%%%%%%%%%%%%%%%%%%%%%%%%%%%%%%%%%%%%%%%%%%%%%%%%%
% \clearpage
% \newpage
% \input{checklist.tex}

\end{document}